\documentclass[runningheads]{llncs}
\usepackage{graphicx}
\usepackage{booktabs}
\usepackage{bbm}
\usepackage{threeparttable}
\usepackage{color,graphicx}
\usepackage{amsmath}
\usepackage[ruled]{algorithm2e}
\usepackage{appendix}
\usepackage{bm}
\usepackage{multirow}
\usepackage{booktabs}
\usepackage{amssymb}
\usepackage{arydshln}
\usepackage{bbm}
\usepackage{threeparttable}
\usepackage{mathrsfs}
\usepackage[ruled]{algorithm2e}
\usepackage{multirow}
\usepackage{booktabs}
\usepackage{amssymb}

\usepackage{bm}
\usepackage{multirow}
\usepackage{booktabs}
\usepackage{amssymb}

\newcommand{\ie}{\textit{i.e.,} } 
\providecommand{\zxhreftb}[1]{Table~\ref{#1}} 
\providecommand{\zxhreffig}[1]{Figure~\ref{#1}} 
\usepackage{changes}
\usepackage{subfigure}
\definechangesauthor[name={Review 1}, color=orange]{R1}
\definechangesauthor[name={Review 2}, color=blue]{R2}
\definechangesauthor[name={Review 1,2}, color=red]{R1+R2}

\usepackage{verbatim}

\begin{document}
\title{ZScribbleSeg: Zen and the Art of Scribble Supervised Medical Image Segmentation}
\titlerunning{ZScribbleSeg}
%

\author{Ke Zhang \and Xiahai Zhuang
\thanks{Xiahai Zhuang is corresponding author. This work was funded by the National Natural Science Foundation of China (Grant No. 61971142 and 62111530195).}}
\authorrunning{K Zhang \& X Zhuang}
%
\institute{School of Data Science, Fudan University, Shanghai\\
\email{zxh@fudan.edu.cn}}
\maketitle              
\begin{abstract}
Curating a large scale fully-annotated dataset can be both labour-intensive and expertise-demanding, especially for medical images. To alleviate this problem, we propose to utilize solely scribble annotations for weakly supervised segmentation. Existing solutions mainly leverage selective losses computed solely on annotated areas and generate pseudo gold standard segmentation by propagating labels to adjacent areas. However, these methods could suffer from the inaccurate and sometimes unrealistic pseudo segmentation due to the insufficient supervision and incomplete shape features. Different from previous efforts, we first investigate the principle of ''good scribble annotations'', which leads to efficient scribble forms via supervision maximization and randomness simulation. Furthermore, we introduce regularization terms to encode the spatial relationship and shape prior, where a new formulation is developed to estimate the mixture ratios of label classes. These ratios are critical in identifying the unlabeled pixels for each class and correcting erroneous predictions, thus the accurate estimation lays the foundation for the incorporation of spatial prior. Finally, we integrate the efficient scribble supervision with the prior into a unified framework, denoted as ZScribbleSeg, and apply the method to multiple scenarios. Leveraging only scribble annotations, ZScribbleSeg set new state-of-the-arts on four segmentation tasks using ACDC, MSCMRseg, MyoPS and PPSS datasets.
\end{abstract}
	In recent years, deep neural networks has demonstrated its potential on various visual tasks~\cite{lecun2015deep}.  
 However, the success of these methods relies on massive annotations, which require assiduous manual efforts.
	For medical imaging, the dense manual labeling can take several hours to annotate just one image for experienced doctors, which is both expensive and expertise-demanding~\cite{zhuang2016multi}.
	Humorous efforts have contributed to the area of training segmentation networks with weaker annotations~\cite{tajbakhsh2020embracing}, including scribbles~\cite{l2016inscribblesup}, bounding boxes~\cite{papandreou2015weakly}, points~\cite{bearman2016s}, and image-level labels~\cite{pathak2014fully}.
	Numerous studies have been reported utilizing only image-level labels~\cite{huang2018weakly,wang2020self,zhang2021affinity,wang2022looking}. These methods mainly rely on large-scale training datasets, and tend to underperform on small medical image datasets.
	On the contrary, scribbles are suitable for labeling nested structures and easy to obtain in practice. 
 Several works have demonstrated their potential on both semantic and medical image segmentation~\cite{khoreva2017simple,koch2017multi,l2016inscribblesup}.
	Therefore, we propose to investigate this specific form of weakly supervised segmentation, which only uses scribble annotations for model training.
	
    Conventionally, scribble annotations are mainly focused on delineating the structure of interests~\cite{9389796}. 
    This can be effective in segmenting \textit{regular structures}, \textit{i.e.,} the targets with fixed shape patterns. 
    Hence, this task is also referred to as \textit{regular structure segmentation}. 
    However, such methods could be challenged when they were applied to portray the irregular targets with heterogeneous distributions, such as pathologies. 
    This is also referred to as \textit{irregular (object) segmentation}, which is particularly challenging for the medical tasks with small training datasets.
    Existing scribble learning approaches mainly aim to reconstruct complete labels from scribbles, and use the generated pseudo labels for model training.
    These works include \textbf{1)} label expansion strategies that assume the pixels with similar features are likely to be in the same category~\cite{ji2019scribble,l2016inscribblesup}, and \textbf{2)} ensemble methods that generate labels by fusing several independent predictions~\cite{luo2022scribble}.
    These methods could be susceptible to the label noises introduced by imprecise segmentation proposals.
    To overcome this issue, Obukhov \emph{et al.} proposed a regularization loss~\cite{obukhov2019gated}, which exploited the similarity between labeled and unlabeled area.
    Adversarial learning approach has also been applied to scribble supervised segmentation~\cite{9389796}, by leveraging shape prior provided by additional full annotations.
    
    Scribble supervised segmentation generally suffers from inadequate supervision and imbalanced label classes.
    This leads to poor results, typically of \textit{under segmentation} of target structures, meaning the volumes of segmented structures tend to be shrunk, as we shall describe in Section~\ref{problems}.
    To address the problem of inadequate supervision, we first investigate the principles of generating ''good scribbles'', as a guidance for designing methodologies to augment supervision, as well as for generating manual annotations.  
    The aim is to model efficient scribbles by maximizing the supervision without increasing annotation efforts.
    Our studies demonstrate that the model training benefit from the randomness of wide range distributed scribbles and larger proportion of annotated areas.
    Inspired by this, we propose to simulate such types of scribble-annotated images as a means of \textit{supervision augmentation}. 
    This can be achieved  via mixup and occlusion operations on existing training images, and the supervision augmentation is coupled with regularization terms penalizing any inconsistency in the segmentation results.  
    
Despite the lack of supervision, the scribble annotations typically have imbalanced annotated label proportions thus biased shape information.
This means the model 
cannot accurately capture the global shape of target structures.
We therefore further propose to correct the problematic prediction using prior-based regularization, particularly from the spatial prior.
This requires the preceding yet critical step of estimating the mixture proportion (ratio) of each label class (referred to as  $\bm\pi$ prior).
We hence propose a new algorithm to compute this $\bm\pi$ prior, 
based on which we develop a spatial loss on the basis
of marginal probability of pixels belonging to certain label
classes and spatial energy.
This spatial loss is a regularization term aimed to correct the shape of segmentation results.
The supervision augmentation and prior-based regularization work in a complementary way, and both contribute to the stable and robust training on a variety of segmentation tasks.


The proposed scribble supervision-based segmentation method, referred to as ZScribbleSeg, extends and generalizes the algorithms in our two preliminary works~\cite{zhang2022cyclemix,zhang2022shapepu}, and has more scientific  significance in the following aspects:
Firstly, we investigate principles of efficient scribble forms to guide the supervision augmentation, which have never be reported to the best of our knowledge.
Secondly, we leverage spatial prior to adjust the predicted probability with computed spatial energy.
Thirdly, we implement a series of extensive experiments on various scenarios, including irregular structure segmentation of medical pathology and visual object segmentation. 
The contributions of this paper are summarized as follows.
    
\begin{itemize}
\item We propose a unified framework for scribble-supervised segmentation by modeling efficient scribbles, and correcting the network prediction with prior regularization, which significantly alleviates the problems of inadequate supervision and imbalanced label classes. 
\item To the best of our knowledge, this is the first work investigating the principles of scribble forms. Motivated by the conclusion that network benefits from larger and randomly distributed annotation, we model efficient scribbles by maximizing supervision and simulating randomness.
\item We propose a novel mechanism to correct the shape of model prediction based on prior regularization, including $\bm{\pi}$ prior, spatial prior, and shape prior.
A new algorithm is introduced to estimate $\bm{\pi}$ prior, based on which we further encode spatial relationship with spatial prior loss.
\item Our approach achieved state-of-the-art performance for weakly-supervised segmentation on regular structures from cardiac anatomical imaging, regular structures from pathology enhanced imaging, irregular objects of medical pathology, and human pose from natural scene.
\end{itemize}

 \begin{figure}[!t]
    \centering
    \includegraphics[width=\linewidth]{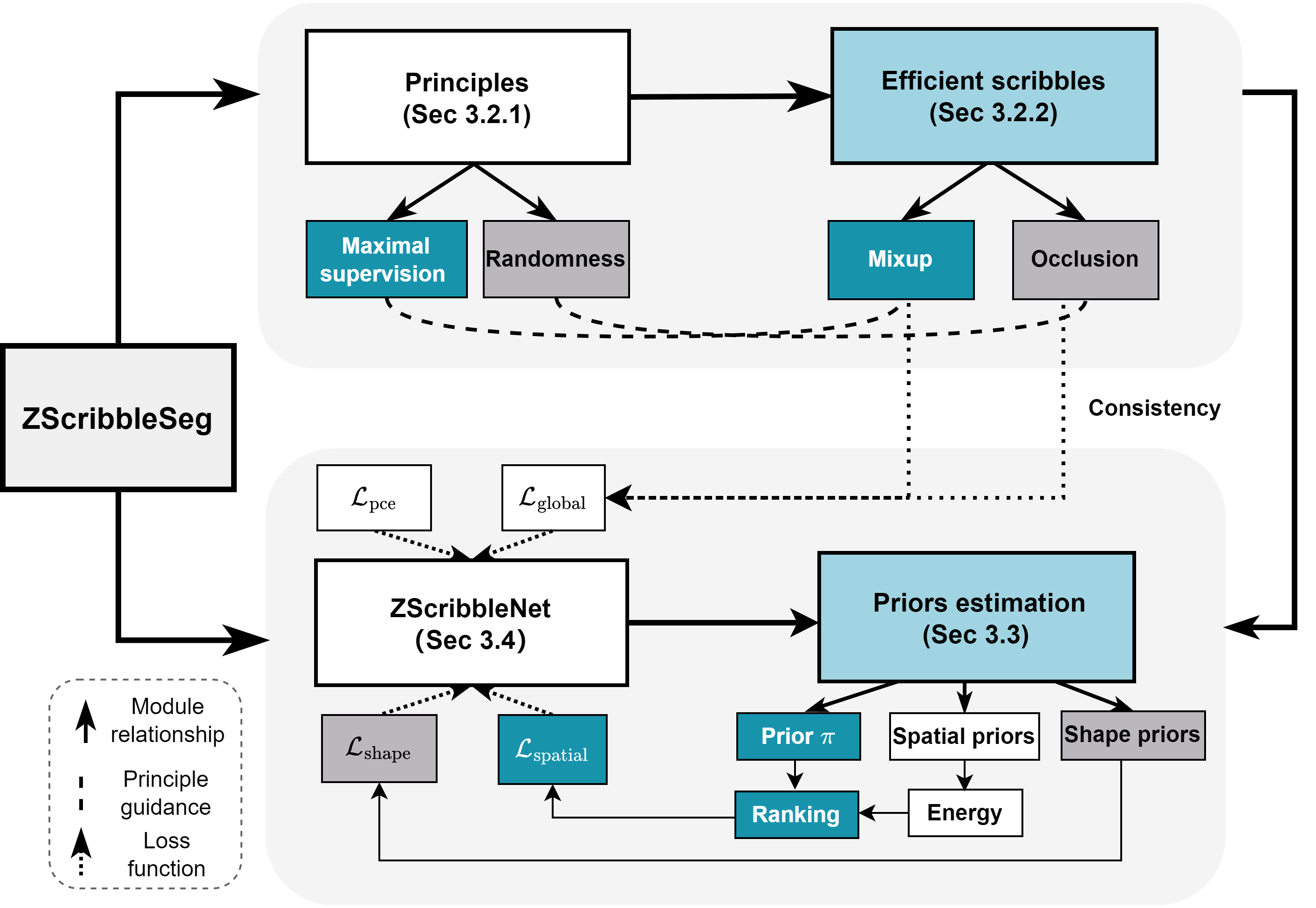}
\caption{Roadmap of the proposed ZScribbleSeg framework.
}
\label{fig:roadmap}
\end{figure}

    The rest of this paper is organized as follows: Section 2 briefly introduces the relevant researches. 
    In section 3, we describe the modeling of efficient scribbles and computation of prior.
    Section 4 presents the results of efficiency, ablation, and validation study.
    Finally, we conclude this work in Section 5.
	
	\section{Related work}
	This section provides a brief review of weakly supervised segmentation methods. Besides, we describe data augmentation strategies and regularization loss functions that closely related to our work.
	
	\subsection{Weakly supervised segmentation}
	Recently, a variety of weakly supervised segmentation strategies have been developed to reduce the manual annotation efforts~\cite{l2016inscribblesup,bearman2016s,papandreou2015weakly,pathak2014fully}. 
	Among them, the scribbles are of particular interest for the application to medical image annotation, given by its advantage in annotating nest structures compared to bounding boxes.
	Current weakly supervised learning methods with image-level annotations mainly generate label seeds with Class Activation Map (CAM)~\cite{zhou2016learning} at first, and then train the network with refined pseudo labels.
	However, the training of CAM requires a large scale of training data labeled with rich visual classes, which is not practical in clinical applications. 
	Therefore, we propose to investigate the scribble supervised segmentation, due to its efficiency and effectiveness in both medical and visual scenarios.

    Scribble is a form of sparse annotation that provides labels for a small subset of pixels in an image~\cite{tajbakhsh2020embracing}.
    Previous approaches mainly calculate losses for annotated pixels.
    One group of works is designed to expand the annotations and reconstruct the full label for network training.
    However, the expansion of labels needs to be achieved through iterative computation, which is particularly time-consuming.
    To alleviate it, several works removed the relabeling process and instead adopted conditional random fields to perform the refinement of segmentation results \cite{chen2017deeplab,Can2018LearningTS,zheng2015conditional,Tang2018OnRL}.
    However, the common issue is the unstable model training caused by noisy pseudo labels. 

    To obtain high-quality pseudo labels and update it throughout the training process, Luo \emph{et al.}~\cite{luo2022scribble} proposed to mix the predictions from dual-branch network as auxiliary pseudo label.
    This approach has achieved promising results on cardiac segmentation, but still susceptible to inaccurate supervisions, especially on more challenging tasks with irregular objects.
    Obukhov \emph{et al.}~\cite{Obukhov2019GatedCL} introduced the Gated CRF loss for unlabeled pixels, which regularizes model training by exploiting the structural similarity between labeled and unlabeled data.
    Other works \cite{9389796,zhang2020accl} included a new module to evaluate the quality of segmentation masks, which encourages the predictions to be realistic, but requiring extra full annotations.
	\subsection{Data augmentation}
    Augmentation methods are investigated to improve the model generalization ability, by synthesizing virtual training examples in the vicinity of the training dataset~\cite{bishop2006pattern}.
    Common strategies include random cropping, rotation, flipping and adding noise~\cite{bishop1995training}.
    Recently, a line of research works have been proposed on Mixup augmentation~\cite{zhang2018mixup,devries2017cutout,yun2019cutmix,kimICML20,kim2021comixup}, which blends two image-label pairs to generate new samples for classification tasks.
    Input Mixup~\cite{zhang2018mixup} was introduced to perform linear interpolation between two images and their labels. 
    Manifold Mixup~\cite{verma2019manifold} applied the Mixup operation to feature space.
    Cutout~\cite{devries2017cutout} randomly occluded a square region of image, and CutMix~\cite{yun2019cutmix} transplanted the occluded area to another image.
    Kim \emph{et al.}~\cite{kimICML20} proposed Puzzle Mix to leverage the saliency and local statistics to facilitate image combination.
    Comixup~\cite{kim2021comixup} extended this concept from two images to multiple images.

    For medical image analysis, Mixup methods have been adopted for image segmentation~\cite{chaitanya2019semi} and object detection tasks~\cite{wang2020focalmix}.
    Although mixup operation may generate unrealistic samples, mixed soft labels can provide rich information and improve the model performance on semi-supervised segmentation~\cite{chaitanya2019semi}.

    \subsection{Regularization losses}
    Neural networks are used to perform pixel-wise image segmentation, typically trained with cross entropy or Dice loss, which computes loss for each pixel independently.
    To predict segmentation coherent in the global sense~\cite{kohl2018probabilistic}, several methods are proposed to regularize the model training.
    Here, we focus on the consistency regularization and $\bm{\pi}$ prior regularization that most relevant to our work.

    The consistency regularization leverages the fact that the perturbed versions of the same image patch should have the consistent segmentation.
    A series of researches have been conducted on consistency regularization~\cite{zhu2017unpaired,laine2016temporal,NIPS2017_68053af2,ouali2020semi}.
    For semi-supervised learning, regularization is applied to the augmented versions of the input image by requiring consistency to obtain stable predictions for unlabeled images~\cite{laine2016temporal,NIPS2017_68053af2,ouali2020semi}.

    The proposed regularization of $\bm{\pi}$ prior is inspired from the binary mixture proportion estimation~\cite{bekker2018estimating,garg2021mixture,ramaswamy2016mixture}, which was originally designed for binary (two-class) positive unlabeled learning~\cite{du2015convex,du2014analysis,NIPS2017_7cce53cf}.
    For multi-class segmentation, the mixture ratios of classes are both imbalanced and inter-dependent, which cannot be solved by existing binary estimation methods.
	        
 \begin{figure*}[t]
    \centering
    \includegraphics[width=\textwidth]{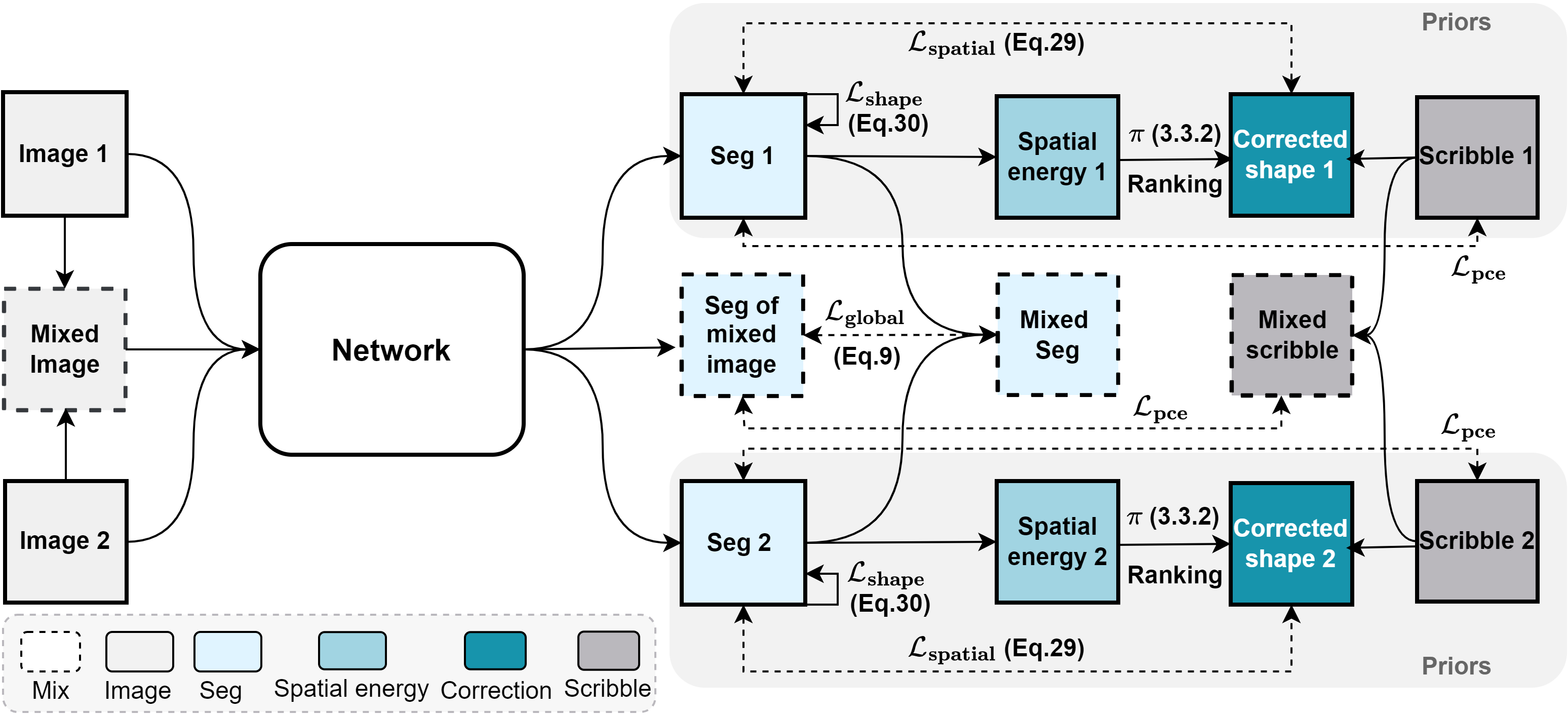}
\caption{
Overview of the training losses for the proposed ZScribbleNet, which consists of modeling of efficient scribbles and computation of priors. 
The scribble modeling includes mixup augmentation, regularized with global consistency ($\mathcal{L}_{\text{global}}$). 
The priors have three, \ie class mixture ratios ($\bm{\pi}$), spatial prior and shape prior, which contribute to spatial prior loss ($\mathcal{L}_{\text{spatial}}$) and shape prior loss ($\mathcal{L}_{\text{shape}}$).
Note that spatial prior loss is complementary with the partial cross entropy loss ($\mathcal{L}_{\text{pce}}$) which is solely calculated for labeled pixels.
}
\label{fig:overview}
\end{figure*}

\section{Method}
\subsection{Overview}
\textbf{Problem Setup:} 
This work investigates the scenario of scribble supervised segmentation, where the training images are solely annotated with a small number of pixels, via scribbles, for each label class.

\noindent\newline\textbf{Strategy:} 
Instead of solely focusing on techniques of weak supervision, we first investigate different forms of scribbles to derive principles of efficient scribbles, \textit{i.e.,} maximal supervision without increasing scribble efforts. 
These principles enable effective and robust model training with minimal annotation cost.
Then, we focus on tackling the major problem of under segmentation, to correct model prediction with prior.

\noindent\newline\textbf{Solution:}
We develop ZScribbleSeg consisting of
(1) modeling efficient scribbles via supervision maximization and randomness simulation.
(2) modeling and computation of prior, including label class proportion prior, spatial prior and shape prior.
(3) integration to develop deep neural network (referred to as ZScribbleNet) having losses of partial cross entropy ($\mathcal{L}_{\text{pce}}$), global consistency $(\mathcal{L}_{\text{global}})$, spatial prior loss ($\mathcal{L}_{\text{spatial}}$), shape regularization $(\mathcal{L}_{\text{shape}})$ and training strategy of supervision augmentation and prior regularization. \zxhreffig{fig:roadmap} presents the roadmap of the proposed framework.

\subsection{Principle and modeling of efficient scribbles}
We investigate the principles of efficient scribbles and derive the objective of maximizing supervision with minimal annotation efforts.
This leads to the proposal of supervision augmentation. In addition, we propose a global consistency loss to penalize the non-equivalence in the augmentation.

\subsubsection{Principles of efficient scribbles}
We shall verify the two principles of achieving efficient scribble annotation in terms of maximal supervision later through the experiments in Section~\ref{sec-exp-scribble}:
\noindent\newline
(1) The large proportion of pixels annotated by scribbles compared with the whole set. 
\noindent\newline
(2) The randomness of distribution of scribbles. This is represented by the random and wide-range annotations. 

Firstly, we are motivated by the knowledge that model training benefits from the finer gradient flow through larger proportion of annotated pixels~\cite{tajbakhsh2020embracing}.
Therefore, we try to increase the annotation proportion with the same effort. 
One natural idea is to simply expand the width of scribbles.
However, this way only increases the label amount in local area, and lacks the ability to enlarge annotation range across the entire image.
    
Secondly, we are inspired by the fact that the imaging data are easier to be restored from random samples of pixels than from down-sampled low-resolution images with regular patterns~\cite{gao2020robust}.
This was due to the fact that the randomly and sparsely distributed samples maintain the global structure of the imaging data, which therefore can be restored with existing low-rank or self-similarity regularization terms. 
By contrast, the regularly down-sampled low-resolution images have evidently reduced tensor ranks, compared with the original high-resolution data, thus lose the global structure information. 
Motivated by this, we assume the features of full segmentation (similarly to the global structure information) can be portrayed (restored) with sparse scribble annotations randomly and widely distributed within the entire dataset. 
With such scribble annotation, the segmentation network can easily learn the global shape prior.
     
Based on the observations described above, we propose to model efficient scribbles by supervision augmentation simulating large annotation proportion and randomness of scribble distribution. 
    
\subsubsection{Modeling via supervision augmentation}
We aim to generate training images with efficient scribbles by maximizing the supervision via mixup operations and achieving the randomness via occlusion operations. 
 This resembles data augmentation, which increases the data diversity and enables robust training.

\noindent\newline\textbf{Search optimal annotation with mixup:} 
    Motivated by the principles of efficient scribble, we first seek the optimal scribble with large annotated ratio, high supervision, and the unchanged local features.
    To achieve that, instead of maximizing the annotations directly, \emph{we aim to maximize the saliency of mixed images}, which measures the sensitivity of model to inputs.
    Given that the annotated area tends to be accompanied with high saliency, maximizing saliency also increases the scribble annotations.

    For two image-scribble pairs $(X_1,Y_1),(X_2,Y_2)$ of dimension $n$, we denote the resulted mixed image-label pair as $(X'_{12},Y'_{12})$.
    The transportation process is defined by:
    \begin{gather}
        X'_{12} = T(X_1, X_2)\ \text{\ and\ }\ Y'_{12} = T(Y_1, Y_2), \label{eq:miximage}\\
        T(X_1, X_2) = (1-\beta)\odot \textstyle\prod_1 X_1+ \beta\odot \textstyle\prod_2 X_2, \label{eq:mixop}
    \end{gather}
    where $T(X_1,X_2)$ represents the transportation process between image $X_1$ and $X_2$; 
    $\scriptstyle\prod_i$ denotes the transportation matrix of size $n\times n$ for image $X_i$; $\beta$ means the mask with value $[0, 1]$ of dimension $n$; $\odot$ is the element-wise multiplication. 
    Then, we aim to maximize the saliency of transportation result over the parameters $\{{\scriptstyle\prod_1}, {\scriptstyle\prod_2},\beta\}$:
    \begin{equation}
    \{{\scriptstyle\prod_1}, {\scriptstyle\prod_2},\beta\} = 
    \underset{{\scriptstyle\prod_1},{\scriptstyle\prod_2}, \beta}{\arg\max}[ (1-\beta)\odot {\scriptstyle\prod_1} M(X_1) + \beta\odot {\scriptstyle\prod_2} M(X_2) ],
    \end{equation}
    where $M(X)$ denotes the saliency map of image $X$, which is obtained by computing the $l_2$ norm of gradient values. We solve this optimization problem based on PuzzleMix~\cite{kimICML20}. 
    To preserve the local statistic features, the optimization objective also includes the image local smoothness, and the mixing weight prior. 
    For details of the optimization objective, we refer readers to PuzzleMix~\cite{kimICML20} and Appendix A of supplementary materials.
    

\begin{figure*}[!t]
    \centering
    \includegraphics[width=1\textwidth]{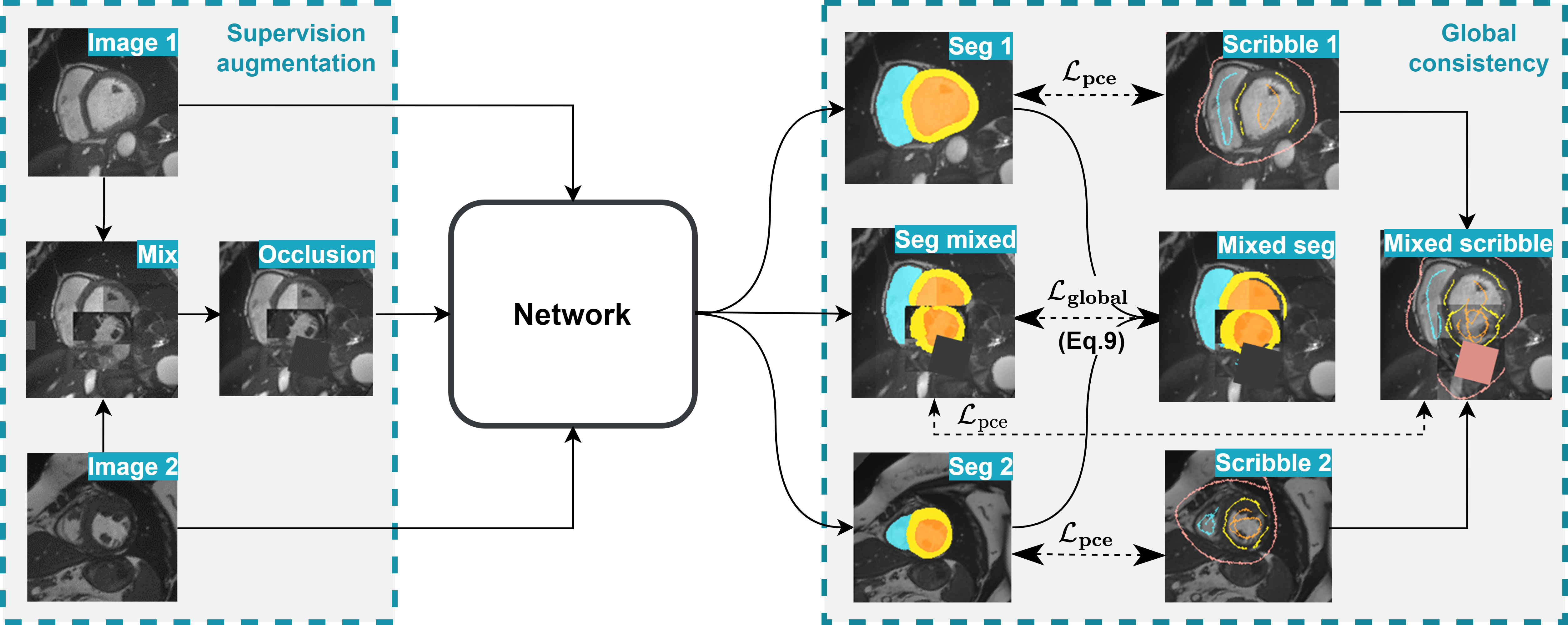}
\caption{Illustration of supervision augmentation and global consistency. Supervision maximization is achieved with the mix augmentation to increase the annotated proportion and data variety. Global consistency requires the segmentation result of mixed image and unmixed image to be consistent.
}
\label{fig:illustration}
\end{figure*}

\noindent\newline\textbf{Introduce randomness via occlusion:}  We propose to simulate randomly distributed scribbles via occlusion. Specifically, one square area of the mixed image is randomly dropped and replaced with the background. 
    Since that the proportion of the background annotated by scribbles tends to be smaller than that of the foreground classes, the occlusion operation alleviates the imbalance problem of class mixture ratios within labeled pixels, and further improves the results of mixture ratio estimation, which will be elaborated in Section~\ref{Sec.estimation}.
   
    We denote the occluded image-label pair as $(X'', Y'')$, which is obtained by: 
    \begin{gather}
        X''_{12} = (1-\mathbbm{1}_b) \odot X'_{12} \\
        Y''_{12} = (1-\mathbbm{1}_b) \odot Y'_{12}
    \end{gather}
    where $\mathbbm{1}_b$ denotes a rectangular mask of size $n\times n$ with value in $[0,1]$. 
    The rectangular mask is randomly rotated to occlude the mixed image, and turns the occluded area into background.
    Following~\cite{yun2019cutmix}, we set the size of rectangular to be $32\times32$.

\noindent\newline\textbf{Global consistency loss:}
    The objective of global consistency regularization is to leverage the mix-invariant property.
    As \zxhreffig{fig:illustration} shows, global consistency requires the same image patch to have consistent segmentation in two scenarios, \textit{i.e.}, the unmixed image and the mixed image.
    Let the segmentation result of image $X$ predicted by network be $\hat{Y} = f(X)$. For the transported image $X'_{12} = T(X_1,X_2)$, the consistency of mixup is formulated as:    
    \begin{equation}
        T(f(X_1), f(X_2)) = f(T(X_1, X_2)),
    \label{3.1eq1}
    \end{equation}
    which requires the segmentation of mixed image to be consistent with the mixed segmentation, after the same transportation process. 
    When applying the occlusion operation, we further have:    
    \begin{equation}
        (1-\mathbbm{1}_b)\odot T(\hat{Y}_1, \hat{Y}_2) 
        = 
        f\left((1-\mathbbm{1}_b)\odot T(X_1, X_2)\right).
    \label{4.1eq2}
    \end{equation}
    Then, we propose to minimize the distance between two sides of Eq.(\ref{4.1eq2}).
    Let $u_{12}=(1-\mathbbm{1}_b)\odot T(\hat{Y}_1, \hat{Y}_2)$ and $v_{12} =f\left((1-\mathbbm{1}_b)\odot T(X_1, X_2)\right)$. 
    The negative cosine similarity $\mathcal{L}_{n}(u_{12}, v_{12})$ is defined as:
    \begin{equation}
        \mathcal{L}_{n}(u_{12}, v_{12}) = -\frac{u\cdot v}{||u_{12}||_2 \cdot ||v_{12}||_2}.
    \end{equation}    
    Taking the symmetrical metric into consideration, we similarly penalize the inconsistency between $u_{21}$ and $v_{21}$. 
    Therefore, the global consistency loss is formulated as:
    \begin{equation}
        \mathcal{L}_{\text{global}} = \frac{1}{2}\left[\mathcal{L}_{n}(u_{12},v_{12}) + \mathcal{L}_{n}(u_{21}, v_{21})\right].
    \label{4.1eq4}
    \end{equation}        
\textbf{Discussion:} Mixup operations could change the shape of target structures, resulting in the unrealistic image. 
     To tackle it, as shown in \zxhreffig{fig:illustration}, we propose to combine the partial cross entropy (PCE) loss for labeled pixels of both mixed and unmixed image, and leverage mix equivalence to preserve shape consistency at global level.
     To further exploit the shape features, we propose to correct the network prediction guided by computed prior, which is described in Section~\ref{sec3.3}.

    \begin{figure}[!t]
        \centering
        \includegraphics[width=1\textwidth]{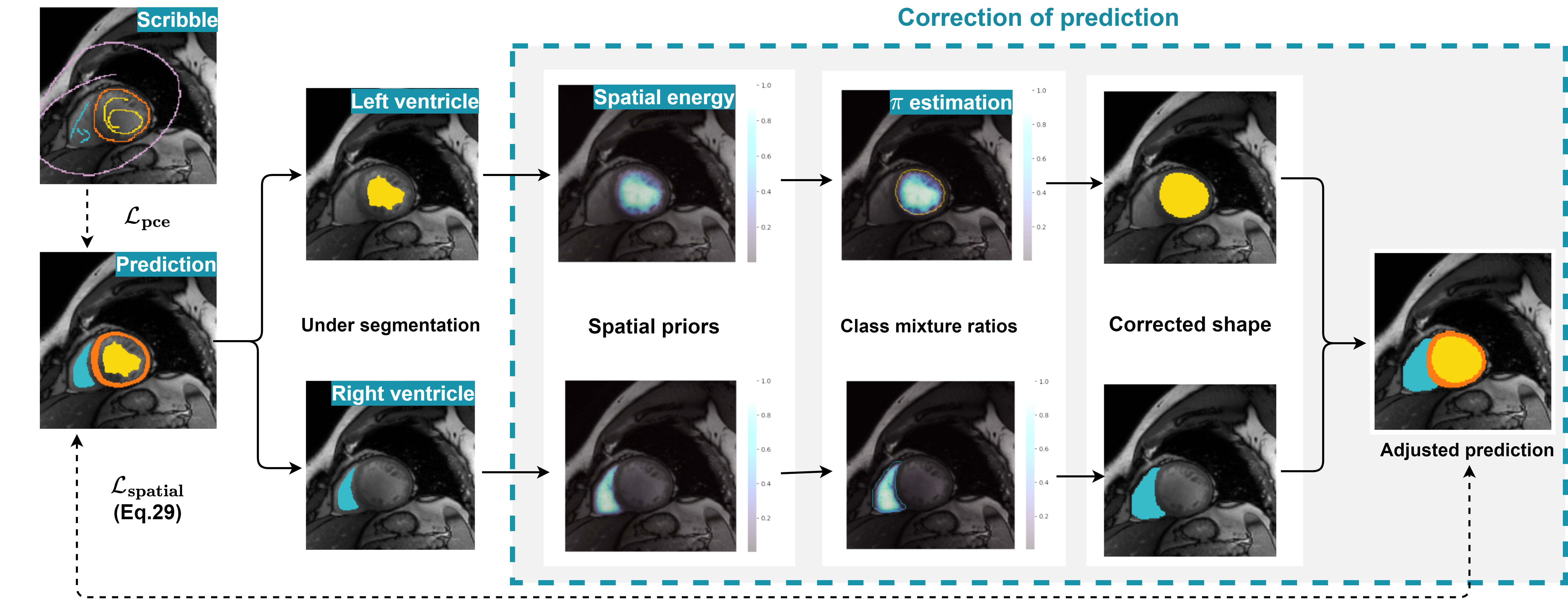}
        \caption{Illustration of spatial prior loss ($\mathcal{L}_{\text{spatial}}$) for correction of prediction, via class mixture ratios ($\bm{\pi}$) and spatial prior (with spatial energy). 
        }
        \label{fig:spatial}
    \end{figure}

\subsection{Modeling and computation of prior}
 \label{sec3.3}
As shown in \zxhreffig{fig:roadmap}, we model class mixture ratios, spatial prior, and shape prior to better capture global shape information and regularize the network training.
As visualized in \zxhreffig{fig:spatial}, we compute the spatial energy to reflect the probabilities of pixels belonging to each class.
We propose a new formulation to estimate critical prior of label class proportions, referred to as $\bm{\pi}$, which guides the correction of erroneous network prediction.

\begin{figure}[t]
    \centering
    \includegraphics[width=0.7\linewidth]{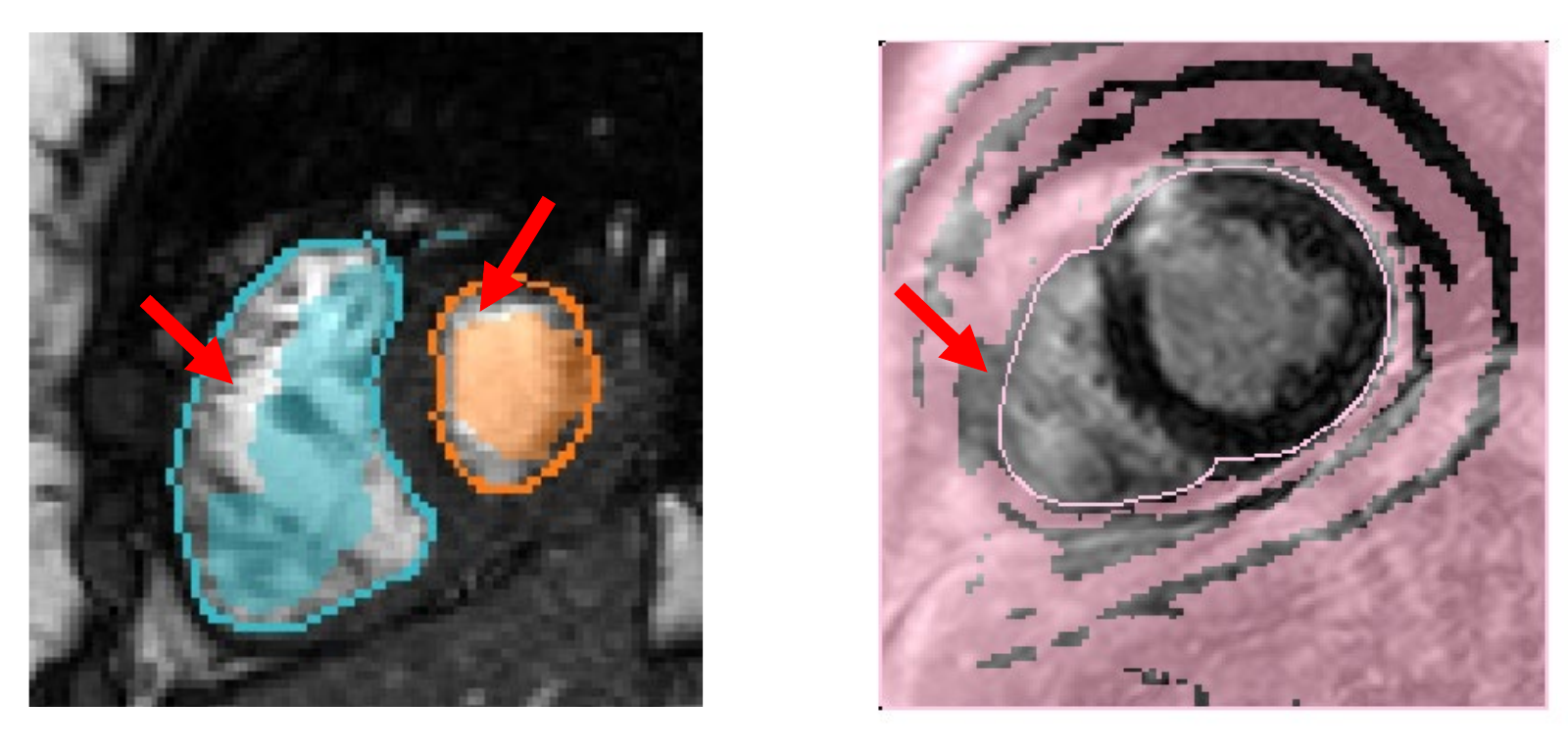}\\
    \makebox[\linewidth][s]{\ (a)\ \ (b)\ }
    \caption{Two examples of under segmentation, pointed by the red arrows: (a) under segmented foreground labels from ACDC segmentation, \ie left ventricle and right ventricle; (b) under segmented background from MyoPS segmentation.
    } \label{fig:underseg}
\end{figure}

\subsubsection{Problems statement}
\label{problems}
The segmentation network trained with scribbles tends to generate under segmentation results of the target structures. 
Considering that the annotated ratio of classes can be imbalanced, the scribble supervised learning also brings challenges to the estimation of class mixture ratios $\bm{\pi}$.

\noindent\newline\textbf{Under segmentation:} 
As shown in \zxhreffig{fig:underseg}, under segmentation refers to the results, where the size of segmented structure is generally smaller than ground truth, a phenomenon caused by the imbalanced annotated proportion and missed shape information. 
To solve the problem, we propose to evaluate $\bm{\pi}$ and spatial prior, which are crucial for the shape refinement.
The accurate estimation of $\bm{\pi}$ can correct the imbalanced label ratios, and enable model to adjust the size of segmentation result.
The computation of spatial prior is able to encode the feature similarity between pixels, and rectify the shape of target structures. 
We encode $\bm{\pi}$ and spatial prior with spatial prior loss, by ranking the spatial energy and select the top $\bm{\pi}$ ratio as the segmentation. 
To estimate $\bm{\pi}$, we start from the imbalanced annotated ratios (referred to as $\bm{a}$) and adapt it from labeled pixels to unlabeled pixels.

Note that the problem of under segmentation can be even worse without the modeling of efficient scribbles.
In the case of manually annotated scribbles, the resulting annotations may be distributed in a non-random pattern due to fixed labeling habits, resulting in the biased label distribution across the whole dataset.
This problem could be alleviated by simulating randomly distributed labels through our proposed supervision augmentation.

\noindent\newline\textbf{Challenges of $\bm{\pi}$ estimation:} The evaluation of class mixture ratios is a critical bottleneck in semi-/ weak-/ non-supervised learning, and serves as the basis of classes identification~\cite{garg2021mixture} and variance reduction~\cite{wu2022minimizing,sakai2017semi}.
However, existing methods are mainly proposed for binary classification, and can not be adapted to multi-class scenario directly.
For segmentation task, the class mixture ratios are both imbalanced and interdependent, leading to the decrease in the performance of previous binary estimation approaches.
Despite the class imbalance problem, the scribble supervised segmentation is also faced with the imbalance of annotated class ratios.
For example, the annotated ratio of the background tends to be much smaller than that of the foreground classes.
The imbalance of annotated ratio further enhances the difficulty of  $\bm{\pi}$ estimation.

\subsubsection{Estimation of class mixture ratios $\bm{\pi}$}
    \label{Sec.estimation}
    To tackle the under segmentation, we propose to estimate the class mixture ratios within unlabeled pixels.
    
    \noindent\newline\textbf{Objective:} We aim to determine $\bm{\pi}$ to maximize the likelihood of observed unlabeled pixels. For $n_u$ unlabeled pixels $\bm{x} =[x_1,x_2,\cdots, x_{n_u}]$ sampled from $p_u(x)$, the likelihood of these unlabeled pixels is formulated as:
    \begin{equation}
    \mathcal{L}(\bm{\pi}) = \prod_{i=1}^{n_u} p_u(x_i)=\prod_{i=1}^{n_u}[\sum_{k=1}^m p_u(x_i|c_k)p_u(c_k)],
    \label{eq_likelihood}
    \end{equation}
    where $p_u(x_i|c_k)$ represents the within-class probability of class $c_k\in\{c_0,\cdots, c_m\}$ for unlabeled pixel $x_i$.
    We assume the within-class probabilities of labeled and unlabeled pixels to be unchanged. 
    Then, we estimate $\bm{\pi} = [p_u(c_1),p_u(c_2),\cdots,p_u(c_m)]$ to maximize the likelihood of unlabeled observations in Eq.(~\ref{eq_likelihood}).

    To maximize the likelihood in Eq.(\ref{eq_likelihood}), we follow the EM algorithm in \cite{latinne2001adjusting,mclachlan2007algorithm} and introduce the unknown variable $\bm{s} = (\bm{s}_1, \bm{s}_2,\cdots,\bm{s}_{n_u})$, where $\bm{s}_i$ is the one-hot vector of dimension $m$ with the i-th value equals 1.
    Then, the likelihood $\mathcal{L}(\bm{\pi}|\bm{x},\bm{s})$ is written as:
    \begin{equation}
    \mathcal{L}(\bm{\pi}|\bm{x},\bm{s}) = \prod_{i=1}^{n_u}\prod_{k=1}^m \left[p_u(x_i|c_k)p_u(c_k)\right]^{s_{ik}}.
    \end{equation}
    The log likelihood $l(\bm{\pi}|\bm{x},\bm{s})$ is derived as:
    \begin{equation}
    \begin{aligned}
    l(\bm{\pi}|\bm{x},\bm{s}) =& \sum_{i=1}^{n_u}\sum_{k=1}^m s_{ik} \log(p_u(x_i|c_k))\\
    &+ \sum_{i=1}^{n_u}\sum_{k=1}^m s_{ik}\log(p_u(c_k))
    \end{aligned}
    \end{equation}
    
    \noindent\newline\textbf{E-step:} The E-step of EM algorithm computes the expected value of $l(\bm{s}|\bm{x},\bm{\pi})$ given the observations $\bm{x}$ and current estimate of $\bm{\pi}^{[t]}$,
    \begin{equation}
    \begin{aligned}
    Q(\bm{\pi}|\bm{x},\bm{\pi}^{[t]}) = &\mathbf{\mathbbm{E}}\left[l(\bm{\pi}|\bm{s},\bm{x})|\bm{x},\bm{\pi}^{[t]} \right]\\
    = &\sum_{i=1}^{n_u}\sum_{k=1}^m \mathbf{\mathbbm{E}}(s_{ik}|x_i,\pi^{[t]}_k)\log(p_u(x_i|c_k)) \\
    &+ \sum_{i=1}^{n_u}\sum_{k=1}^m\mathbf{\mathbbm{E}}(s_{ik}|x_i,\pi^{[t]}_k)\log(p_u(c_k)),\\
    \end{aligned}
    \label{E_objective}
    \end{equation}
    where $\mathbf{\mathbbm{E}}(s_{ik}|x_i,\pi^{[t]}_k)$ is represented as:
    \begin{equation}
    \mathbf{\mathbbm{E}}(s_{ik}|x_i,\pi^{[t]}_k)= p(s_{ik}=1|x_i,\pi^{[t]}_k) = p^{[t]}_u(c_k|x_i)
    \end{equation}
    
    \noindent\newline\textbf{Estimation of $\bm{p^{[t]}_u(c_k|x_i)}$:}  
    To solve the current estimate of $p^{[t]}_u(c_k|x_i)$, we aim to adapt the posteriori probability from labeled pixels to unlabeled pixels. 
    For labeled pixels, the posteriori probability $p_l(c_k|x_i)$ is estimated by the model prediction.
    For class $c_k$ and pixel $x_i$, Based on our assumption that the within-class probabilities of labeled and unlabeled pixels are same, we have 
    \begin{equation}
    p_u(x_i|c_k)=p_l(x_i|c_k),
    \label{eq:assumption}
    \end{equation}
    Based on Bayes' theorem, the within-class probabilities of labeled pixel $ p_l(x_i|c_k)$ and unlabeled pixel $p_u(x_i|c_k)$ are written as:
    \begin{equation}
    \hat{p}_l(x_i|c_k)=\frac{\hat{p}_l(c_k|x_i)p(x_i)}{\hat{p}_l(c_k)}
    \label{eq_bayes_l}
    \end{equation}
    \begin{equation}
    \hat{p}_u(x_i|c_k) =\frac{\hat{p}_u(c_k|x_i)\hat{p}_u(x_i)}{\hat{p}_u(c_k)}\\
    \label{eq_bayes_u}
    \end{equation}
    By substituting $\hat{p}_u(x_i|c_k)$ in Eq.(\ref{eq_bayes_u}) and $\hat{p}_l(x_i|c_k)$ in Eq.(\ref{eq_bayes_l}) into Eq.(\ref{eq:assumption}), we adapt the within-class probabilities from labeled pixels to unlabeled pixels as follows:
    \begin{equation}
        \hat{p}_u(c_k|x_i) = \frac{\hat{p}_l(x_i)}{\hat{p}_u(x_i)}\cdot\frac{\hat{p}_u(c_k)}{\hat{p}_l(c_k)}\hat{p}_l(c_k|x_i).
        \label{eq3}
    \end{equation}
   For binary estimation, the mixture ratio is independently estimated for each class, which does not leverage the inter-relationship between classes. 
   For multi-class segmentation, we naturally utilize the condition that the sum of the probabilities of all classes equals to 1, \emph{i.e.}, 
   \begin{equation}
   \sum_{k=0}^m \hat{p}_u(c_k|x_i) = 1.
   \label{multi}
   \end{equation}
   By combing Eq.(\ref{eq3}) and Eq.(\ref{multi}), one can obtain:
    \begin{equation}
        1 = \frac{\hat{p}_l(x_i)}{\hat{p}_u(x_i)}\sum_{k=0}^m\frac{\hat{p}_u(c_k)}{\hat{p}_l(c_k)}\hat{p}_l(c_k|x_i).
    \end{equation}
    Then, $\hat{p}_l(x_i)/\hat{p}_u(x_i)$ is represented as:
    \begin{equation}
        \frac{\hat{p}_l(x_i)}{\hat{p}_u(x_i)} =\left[\sum_{k=0}^m [\hat{p}_u(c_k)\hat{p}_l(c_k|x_i)/\hat{p}_l(c_k)]\right]^{-1}.
    \end{equation}
    By substituting $\hat{p}_l(x_i)/\hat{p}_u(x_i)$ into Eq.~(\ref{eq3}), we can obtain the formulation of $\hat{p}_u(c_k|x_i)$ as follows:
    \begin{equation}
        \hat{p}_u(c_k|x_i) = \frac{\hat{p}_u(c_k)\hat{p}_l(c_k|x_i)/\hat{p}_l(c_k)}{\sum_{k=0}^m[\hat{p}_u(c_k)\hat{p}_l(c_k|x_i)/\hat{p}_l(c_k)]}.
    \end{equation}

    Therefore, the current estimate of posteriori probability $\hat{p}_u(c_k|x_i)$ is written as:
    \begin{equation}
        \hat{p}_u^t(c_k|x_i) = \frac{\pi^t_k\hat{p}_l(c_k|x_i)/\hat{p}_l(c_k)}{\sum_{k=0}^m[\pi^t_k\hat{p}_l(c_k|x_i)/\hat{p}_l(c_k)]},
    \end{equation}
    where $\hat{p}_l(c_k)$ is empirically evaluated by the class frequency within labeled pixels, \textit{i.e.}, $\hat{p}_l(c_k) = n^k_l/n_l$.

    \noindent\newline\textbf{M-step:} The M-step maximizes $Q(\bm{\pi},\bm{\pi}^{[t]})$ in Eq.(\ref{E_objective}), \emph{i.e.},
    \begin{equation}
    \bm{\pi}^{[t+1]} := \underset{\bm{\pi}}{\arg\max}\ Q(\bm{\pi}|\bm{x},\bm{\pi}^{[t]}) 
    \end{equation}
    We empirically solve the $\pi_k^{t+1}$ as:
    \begin{equation}
    \pi_k^{[t+1]} = \frac{1}{n_u} \sum_{i=1}^{n_u}p^{[t]}_u(c_k|x_i)
    \label{M_objective}
    \end{equation}
    
    The $\pi_k^{[t]}$ is initialized with the class frequency within labeled pixels $\bm{a}$, with $a_k = \frac{n^k_l}{n_l}$.
    Then, the E-step of Eq.(\ref{E_objective}) and M-step of Eq.(\ref{M_objective}) is repeated until the estimation of $\bm{\pi}$ converges. 
    The posteriori probability $\hat{p}_u(c_k|x_i)$ and priori probability $ \hat{p}_u(c_k)$ are re-estimated in each iteration.

    \noindent\newline\textbf{Discussion:} There are two conditions of the proposed algorithm. Firstly, we assume the within-class probabilities of labeled and unlabeled pixels be the same, which means the labeled pixels should be randomly sampled based on classes. 
    Secondly, $\bm{\pi}$ is initiated with the class frequency of labeled pixels $\bm{a}$.
    Since that the annotated ratio of background is smaller than that of the foreground classes, the priori probabilities of foreground classes within unlabeled pixels tend to be over-estimated.
    The first problem can be tackled by modeling the efficient scribbles, to achieve the random distribution of annotations. 
    For the second problem, by randomly occluding the image and replace the occluded area with background, we are able to increase the ratio of background and alleviate this problem to some extent. Furthermore, we propose to address it with the marginal probability maximization, which will be explained in Section~\ref{marginal}.
    \subsubsection{Computation of spatial energy}
    Given the estimated class mixture ratios, we aim to identify the unlabeled pixels by determining the probability of pixels belonging to each class.
    Instead of using model predictions directly, we further encode the spatial relationship to compensate the inaccurate results generated by segmentation network.
    Inspired by \cite{Obukhov2019GatedCL}, we estimate the spatial energy of unlabeled pixels with energy term in a dense setting.

	Firstly, we use Gaussian kernels $G_{ij}$ to measure the distance between pixels at position $i$ and $j$ as:
	\begin{equation}
	   G_{ij}  =\exp\left\{-\frac{(p_i-p_j)^2}{2\sigma_p^2}-\frac{(o_i-o_j)^2}{2\sigma_o^2}\right\},
	\end{equation}
	where $p_i$ represents the position of pixel $x_i$; $o_i$ denotes the color feature; $\sigma_p$ and $\sigma_o$ are the bandwidth parameters for position and color information, respectively. The shallow features like color and position are specific to the pixel and do not rely on the network prediction. Then, the energy term $\phi_{ij}$ leveraging prediction $\hat{y}$ is formulated as:
	\begin{equation}
	    \phi_{ij}(\hat{y}) = G_{ij}\hat{y}_i\hat{y}_j,
	\end{equation}
	which denotes the pairwise relationship between two pixels. This energy term connects every pixels with each other within one image. Based on $\phi_{i,j}$, we define the element of spatial energy $\Phi$ in a dense setting, \emph{i.e.},
	\begin{equation}
	    \Phi_i(\hat{y}) = \sum_{j\in\Omega_i} \phi_{ij}(\hat{y}),
	    \label{3.3eq11}
	\end{equation}
	where $\Omega_i = \{\text{Pos}(i)-\text{Pos}(j)\leq r\}$, means the neighborhood window of radius $r$. Instead of taking the total energy as the regularization loss as~\cite{Obukhov2019GatedCL}, we consider $\Phi$ as the spatial energy to reflect the relative probability of pixels belonging to each class. 

\subsubsection{Spatial prior and shape prior losses}
    \label{marginal}
\textbf{Spatial prior loss} is computed by ranking the spatial energy and selecting the top $\bm{\pi}$ proportion of pixels as the segmentation. 
    Considering that adjusting multiple structures directly can be challenging, we instead separate each foreground class from the others, and then tackle the individual structure. 
    Given that the mixture ratios of foreground classes tend to be over-estimated, we instead leverage the accurate negative pixels filtered by estimated mixture ratios, and maximize the marginal probability of these pixels belonging to other classes.

    Firstly, by ranking the spatial energy and applying the mixture ratio of each class, we are able to distinguish negative pixels from unlabeled pixels. 
    For foreground class $c_k$, we rank the unlabeled pixels according to the spatial energy $\Phi^{k}$ of class $c_k$ in Eq.~(\ref{3.3eq11}). 
    Given the estimated mixture ratio $\pi_k$, we set pixels in the top  $\pi_k$ proportion to be positive samples $\Omega_k$
    Correspondingly, the remaining pixels are taken as negative pixels, denoted as $\bar{\Omega}_k$.
    Taking over-estimated $\pi_k$ into account, we believe the set of negative pixels $\bar{\Omega}_k$ is more accurate than $\Omega_k$.

    Secondly, we design the spatial prior loss ($\mathcal{L}_{\text{spatial}}$) based on maximal marginal probability of negative samples $\bar{\Omega}_k$ belonging to other classes.
    For each class $c_k$, we take it as foreground and fuse other classes except $c_k$ into background. 
    The fused class is denoted as $\bar{c}_k$.
    For pixel $x_i$ in $\bar{\Omega}_k$, its marginal probability belonging to $\bar{c}_k$ equals the sum of probabilities of the fused classes, \emph{i.e.}, $\hat{p}(\bar{c}_k|x_i, x_i \in \bar{\Omega}_k) = \sum_{k'=1}^m [\mathbbm{1}_{[k'\neq k]}\hat{p}(c_k|x_i)]$.
    To maximize the marginal probability of negative pixel $x_i$ belonging to $\bar{c}_k$, we formulate the spatial prior loss as:
    \begin{equation}
        \mathcal{L}_{\text{spatial}} =-\sum_{k=1}^m \sum_{x_i\in \bar{\Omega}_k}\log(\hat{p}\left(\bar{c}_k|x_i)\right).
        \label{eq19}
    \end{equation}
    
\noindent\textbf{Shape prior loss} is developed to regularize inter-connected structures in the segmentation results.
This loss is adopted to further reduce noise and smooth boundary.
It requires the model prediction to be consistent with its maximum connected area, and minimizes their cross entropy loss, \emph{i.e.},
    \begin{equation}
    \small
        \mathcal{L}_{\text{shape}} = -\sum_{k \in \varPsi}F(\hat{Y}_k)\log(\hat{Y}_k),
    \label{4.2eq1}
    \end{equation}
    where $\varPsi$ is the set of label classes with inter-connected structures; $F(\cdot)$ denotes the morphological function, and outputs the largest inter-connected area of input label.

\begin{table*}[!thb] 
\caption{Efficiency analysis of scribble forms for regular structure segmentation of cardiac ventricles (ACDC dataset) and irregular segmentation of myocardial pathology (MyoPS dataset). Here,
$N_\text{scribble}$ and $N_\text{pix}$ respectively denote the number of manual draws to generate scribble annotations and number of annotated pixels, which indicate annotation efforts; 
$k$ is the number of manual draws (scribbles) and $n$ is the given threshold of annotation efforts, where $k<<n$. 
Segmentation results are evaluated on test set and reported in Dice scores.
}\label{tab1}
\centering
\resizebox{\textwidth}{!}{
\begin{tabular}{ccc cccc ccc}
\hline  \multirow{2}{*}[-4pt]{Methods}&\multirow{2}{*}[-4pt]{$N_\text{scribble}$}&\multirow{2}{*}[-4pt]{$N_\text{pix}$}&\multicolumn{4}{c|}{Structural segmentation}&\multicolumn{3}{c}{Irregular segmentation}\\
\cmidrule(lr){4-7}\cmidrule(lr){8-10}
&&&LV & MYO & RV & \multicolumn{1}{c|}{Avg} & Scar & Edema &\multicolumn{1}{c}{Avg} \\
\hline

\multicolumn{1}{c|}{Points}&n&\multicolumn{1}{c|}{n}&\textbf{.876$\pm$.134}&\textbf{.801$\pm$.089}&\textbf{.858$\pm$.081}&\multicolumn{1}{c|}{\textbf{.845$\pm$.107}}&
\textbf{.551$\pm$.246}&\textbf{.638$\pm$.115}&\multicolumn{1}{c}{\textbf{.595$\pm$.194}}\\

\multicolumn{1}{c|}{Skeleton}&k&\multicolumn{1}{c|}{n}&.805$\pm$.145&.737$\pm$.095&.769$\pm$.128&\multicolumn{1}{c|}{.770$\pm$.126}& 
.504$\pm$.213&.057$\pm$.022&\multicolumn{1}{c}{.281$\pm$.271} \\

\multicolumn{1}{c|}{Random walk}&k&\multicolumn{1}{c|}{n}&.798$\pm$.173&.698$\pm$.153&.753$\pm$.157&\multicolumn{1}{c|}{.744$\pm$.165}&
.516$\pm$.284&.529$\pm$.123&\multicolumn{1}{c}{.522$\pm$.184}\\

\multicolumn{1}{c|}{DirRandomWork}&k&\multicolumn{1}{c|}{n}&\underline{.844$\pm$.143}&\underline{.755$\pm$.102}&\underline{.798$\pm$.173}&\multicolumn{1}{c|}{\underline{.799$\pm$.146}}&
\underline{.539$\pm$.217}&\underline{.637$\pm$.108}&\multicolumn{1}{c}{\underline{.588$\pm$.176}}\\
  
\hline
\end{tabular}}
\end{table*} 

\begin{figure*}[t]
\centering
\subfigure[]{
\centering
\includegraphics[width=0.45\textwidth]{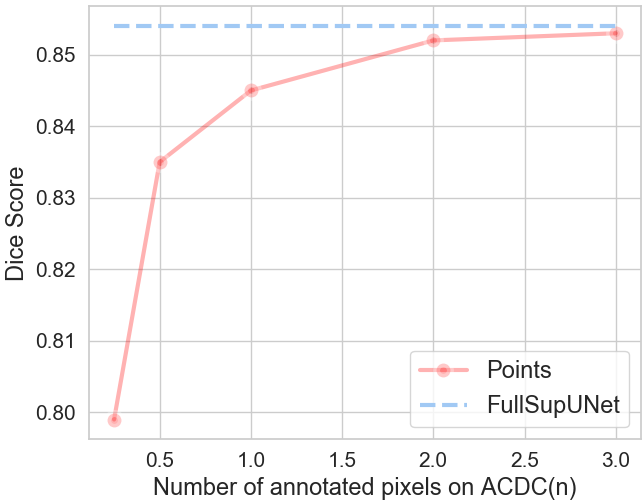}}
\subfigure[]{
\centering
\includegraphics[width=0.45\textwidth]{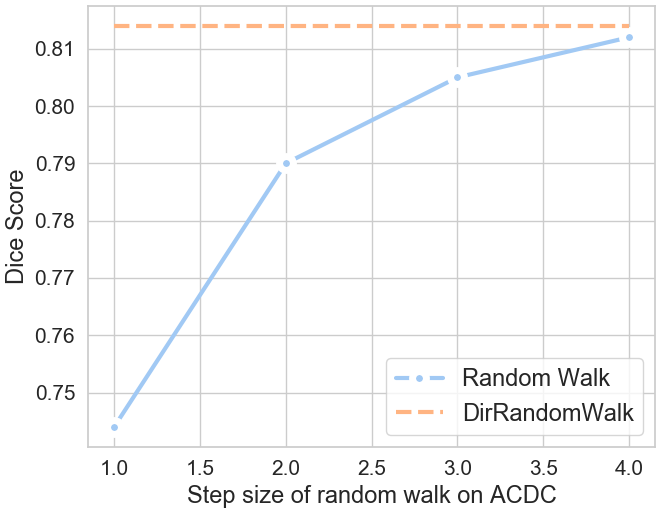}}
\subfigure[]{
\centering
\includegraphics[width=0.45\textwidth]{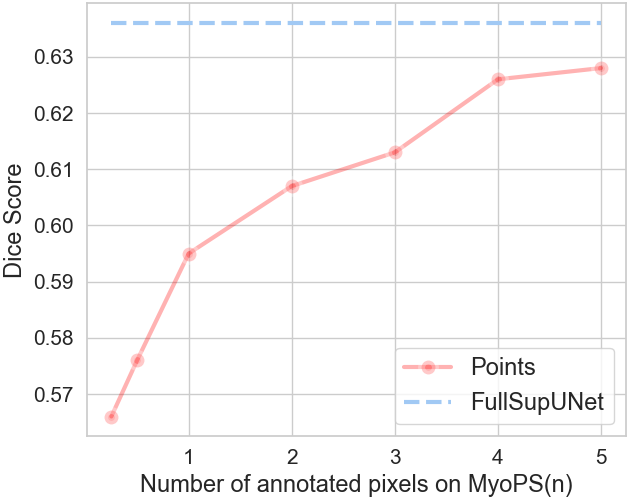}}
\subfigure[]{
\centering
\includegraphics[width=0.45\textwidth]{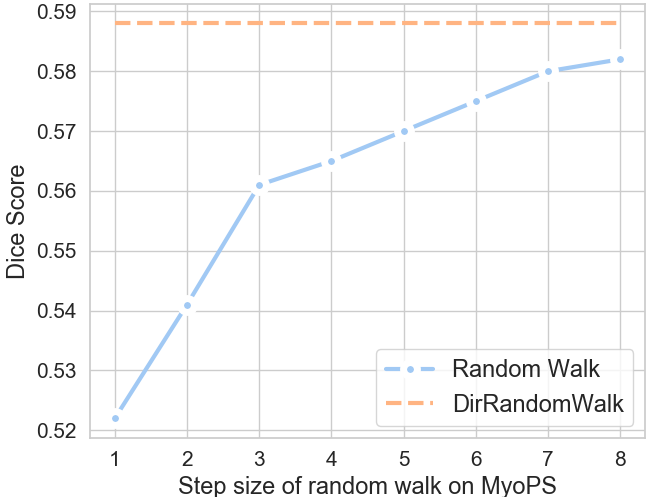}}
\caption{Performance of segmentation networks trained by the Points scribble form with different number of pixels $N_{pix}$, with comparisons to fully supervised models (FullSupUNet): (a) and (c) visualize Dice scores with respect to different $N_{pix}$ on ACDC and MyoPS, respectively. 
The performance of models trained by the Random walk form, with increasing step length $l$, compared with models trained by DirRandWalk:  (b) and (d) show the Dice scores of segmentation on ACDC and MyoPS, respectively, given $N_{pix}=n$.}
\label{fig:exp:scribbleform}\end{figure*}
 

\subsection{ZScribbleNet}
ZScribbleSeg is achieved via a deep neural network referred to as ZScribbleNet.
ZScribbleNet does not depend on any particular network architecture, and can be directly applied to any CNN backbone.
For all experiments, we adopt the variant of UNet \cite{baumgartner2017exploration} as the backbone of segmentation network.
As \zxhreffig{fig:overview} shows,
two images are mixed together to perform the supervision augmentation.
Then, our ZScribbleNet takes the mixed images and unmixed images as the input, and output their segmentation results.

For model training, images and their scribble annotations are sampled to estimate the training objective ($\mathcal{L}$), which is formulated as:
\begin{equation}
    \mathcal{L} = \mathcal{L}_{\text{pce}}+\underbrace{\lambda_1\mathcal{L}_{\text{global}}+\lambda_2\mathcal{L}_{\text{spatial}} +\lambda_3\mathcal{L}_{\text{shape}}}_{\text{unsup}},
\label{eq:final}
\end{equation}
where $\mathcal{L}_{\text{pce}}$ is the partial cross entropy loss calculated for annotated pixels in unmixed image and mixed image; the global consistency loss $\mathcal{L}_{\text{global}}$ in Eq.(\ref{4.1eq4})  requires the mix equivalence for the supervision augmentation; spatial prior loss $\mathcal{L}_{\text{spatial}}$ in Eq.(\ref{eq19}) encodes the $\bm{\pi}$ prior and spatial prior; shape regularization loss $\mathcal{L}_{\text{shape}}$ in Eq.(\ref{4.2eq1}) leverages shape prior; $\lambda_{1},\lambda_2,\lambda_3$ are hyper-parameters to leverage the relative importance of different loss components.

In the training phase, We warmly started training the networks with partial cross entropy loss $\mathcal{L}_{\text{pce}}$, global consistency loss $\mathcal{L}_{\text{global}}$, and shape regularization loss $\mathcal{L}_{\text{shape}}$ for 100 epochs, and then invoked the spatial loss $\mathcal{L}_{\text{spatial}}$. 
In the testing phase, the trained network predicted the segmentation results of input image directly.
    
\section{Experiments and Results}
We first investigated a variety of scribble forms, and analyzed the principles of efficient scribbles in Section~\ref{sec-exp-scribble}. 
Then, we performed ablation study to the proposed ZScribbleSeg in Section~\ref{sec-exp-ablation}.
Finally, we demonstrated the performance of ZScribbleSeg with comparisons to other state-of-the-art methods in various segmentation tasks using four open datasets in Section~\ref{sec-exp-sota}.

\subsection{Materials}\label{sec-exp-materials}

\subsubsection{Tasks and datasets}
Our validation included four segmentation tasks, including (1) regular structure segmentation of cardiac ventricles from anatomical imaging using ACDC dataset, (2) regular structure segmentation from pathology enhanced imaging with a smaller training size using MSCMRseg dataset, (3) irregular object segmentation of myocardial pathology from multi-modality imaging using MyoPS dataset, and human pose segmentation from natural scene images using PPSS dataset.

\textbf{ACDC} dataset was from 
the MICCAI’17 Automatic Cardiac Diagnosis Challenge~\cite{8360453}. 
This dataset consists of short-axis cardiac images using anatomical MRI sequence (BSSFP) from 100 patients, with gold standard segmentation of cardiac ventricular structures, including left ventricle blood cavity (LV), left ventricle myocardium (MYO), and right ventricle blood cavity (RV).
For experiments, we randomly divided the 100 subjects into a training set of 70 subjects, a validation set of 15  subjects (particularly for ablation study), and a test set of 15  subjects.

\textbf{MSCMRseg} was from the MICCAI’19 Multi-sequence Cardiac MR Segmentation Challenge~\cite{8458220,Zhuang2016MultivariateMM}, consisting of images from 45 patients with cardiomyopathy and the gold standard segmentation of LV, MYO and RV. 
We employed the 45 images of late gadolinium enhanced (LGE) MRI to evaluate the segmentation of ventricle structures.
Following \cite{yue2019cardiac}, we divided the 45 images into three sets of 25 (training), 5 (validation), and 15 (test) images for all experiments.
Note that this structure segmentation is more challenging than that on ACDC due to its smaller training set and pathology enhanced images.

\textbf{MyoPS} dataset was from MICCAI'20 Myocardial pathology segmentation Challenge~\cite{li2022myops}, consisting of paired images of BSSFP, LGE and T2 cardiac MRI from 45 patients. 
The task was to segment the myocardial pathologies, including scar and edema, which do not have regular shape or structure thus their segmentation represents a different task to the regular structure segmentation. 
Following the benchmark study~\cite{li2022myops}, we split the data into 20 pairs of training set, 5 pairs of validation set and 20 pairs of test set.

\textbf{PPSS} refers to the Pedestrian Parsing on Surveillance Scenes (PPSS) dataset~\cite{luo2013pedestrian}. 
We employed the task of human pose segmentation to validate the generalizability of models on natural scene images.
PPSS is a large scale human parsing dataset including 3673 annotated samples of 171 surveillance videos.  
The ground truth segmentation of eight classes including hair, face, upper clothes, arms, lower clothes, legs, shoes, and background were provided.
We used the first 100 surveillance scenes for training and the remaining 71 videos for test.



\subsubsection{Evaluation metrics} 
For experiments on ACDC, MSCMRseg and MyoPS datasets, we reported the Dice score and Hausdorff Distance (HD) on each foreground class separately following the practice of medical image segmentation.     
On PPSS dataset, we measured the multi-class Dice scores  following~\cite{9389796}, where Dice= $\frac{2|\hat{\bf{y}}\bf{y}|}{|\hat{\bf{y}}|+|\bf{y}|}$, and $\hat{\bf{y}}$ and $\bf{y}$ denote the multi-channel prediction and ground truth label, respectively.

\subsubsection{Pre-processing and implementation}

The two dimensional slices from ACDC and MSCMR datasets were of different resolutions. Hence, we first re-sampled all images into a fixed resolution of $1.37\times1.37$~mm and then extracted the central patch of size $212\times212$ for experiments.
For MyoPS, we took the paired slices of BSSFP, LGE, and T2 CMR and cropped their center patches of size $192\times192$ for experiments.
We normalized the intensity of these medical images to be zero mean and unit variance.
For PPSS dataset, we first re-sampled all images into the same resolution, and then padded the images to the size of $160\times160$. The intensities of images were normalized to a range between 0 and 1.

For random occlusion, a square area of $32\times 32$ was randomly occluded for each image.
For the estimation of spatial energy, We adopted Gaussian kernels with color bandwidth $\sigma_{o} = 0.1$, position bandwidth $\sigma_p = 6$, and kernel radius $r=5$.
The hyper-parameters $\lambda_1$, $\lambda_2$, $\lambda_3$ in Eq.~(\ref{eq:final}) were empirically set to be $0.05$, $1$, and $1$, respectively.

All models were trained with a batch size of 4, learning rate of 1e$^{-4}$, and augmentation of flipping and random rotation.    
We implemented our models with Pytorch. All models were trained on one NVIDIA 3090Ti 24GB GPU for 1000 epochs.

\begin{table*}[!thb]
\caption{Results in Dice scores and Hausdorff Distance (HD) of the ablation study using ACDC dataset, where the models were evaluated on the validation set. Note that model~\#6 is  ZScribbleSeg.
\textbf{Bold} denotes the best result, and \underline{underline} indicates the best but one in each category.}\label{tab3}
\centering
\resizebox{1\textwidth}{!}{
\begin{tabular}{|c|ccccc|cccc|}
\hline 
Results in Dice&$\mathcal{L}_{\text{pce}}$ & Efficiency&$\mathcal{L}_{\text{shape}}$&$\mathcal{L}_{\text{global}}$&$\mathcal{L}_{\text{spatial}}$& LV & MYO & RV &Avg\\
\hline
model~{\#1}&$\checkmark$&$\times$&$\times$&$\times$&\multicolumn{1}{c|}{$\times$}&.863$\pm$.089&.804$\pm$.063&.774$\pm$.150&.813$\pm$.112\\
model~{\#2}&$\checkmark$&$\checkmark$&$\times$&$\times$&\multicolumn{1}{c|}{$\times$}&.870$\pm$.100&.833$\pm$.063&.843$\pm$.076&.848$\pm$.082\\
model~{\#3}&$\checkmark$&$\times$&$\checkmark$&$\times$&\multicolumn{1}{c|}{$\times$}&.915$\pm$.068&.871$\pm$.056&.871$\pm$.058&.886$\pm$.064\\
model~{\#4}&$\checkmark$&$\checkmark$&$\times$&$\checkmark$&\multicolumn{1}{c|}{$\times$}&.920$\pm$.064&.868$\pm$.051&.886$\pm$.051&.891$\pm$.059\\
model~{\#5}&$\checkmark$&$\times$&$\times$&$\times$&\multicolumn{1}{c|}{$\checkmark$}&\underline{.923$\pm$.078}&\underline{.869$\pm$.051}&\underline{.889$\pm$.056}&\underline{.894$\pm$.066}\\
model~{\#6}&$\checkmark$&$\checkmark$&$\checkmark$&$\checkmark$&\multicolumn{1}{c|}{$\checkmark$}&\textbf{.929$\pm$.057}&\textbf{.876$\pm$.051}&\textbf{.892$\pm$.049}&\textbf{.899$\pm$.056}\\
\hline
\hline
Results in HD (mm)&$\mathcal{L}_{\text{pce}}$ & Efficiency&$\mathcal{L}_{\text{shape}}$&$\mathcal{L}_{\text{global}}$&$\mathcal{L}_{\text{spatial}}$& LV & MYO & RV &Avg\\\hline
model~{\#1}&$\checkmark$&$\times$&$\times$&$\times$&\multicolumn{1}{c|}{$\times$}&81.86$\pm$40.40&65.97$\pm$33.62&60.91$\pm$44.62&69.58$\pm$40.37\\
model~{\#2}&$\checkmark$&$\checkmark$&$\times$&$\times$&\multicolumn{1}{c|}{$\times$}&119.78$\pm$19.14&23.90$\pm$17.32&52.38$\pm$23.40&65.35$\pm$45.06\\
model~{\#3}&$\checkmark$&$\times$&$\checkmark$&$\times$&\multicolumn{1}{c|}{$\times$}&\textbf{4.45$\pm$5.39}&\underline{15.24$\pm$23.90}&25.78$\pm$22.44&\underline{15.16$\pm$20.89}\\
 	model~{\#4}&$\checkmark$&$\checkmark$&$\times$&$\checkmark$&\multicolumn{1}{c|}{$\times$}&12.12$\pm$18.26&29.41$\pm$24.56&16.97$\pm$15.62&19.50$\pm$20.94\\
model~{\#5}&$\checkmark$&$\times$&$\times$&$\times$&\multicolumn{1}{c|}{$\checkmark$}&28.95$\pm$36.57&44.77$\pm$34.69&\textbf{7.51$\pm$5.34}&27.08$\pm$32.76\\
model~{\#6}&$\checkmark$&$\checkmark$&$\checkmark$&$\checkmark$&\multicolumn{1}{c|}{$\checkmark$}&\underline{6.09$\pm$8.53}&\textbf{11.14$\pm$14.53}&\underline{8.86$\pm$5.88}&\textbf{8.70$\pm$10.40}\\
\hline
\end{tabular}}
\end{table*}  

\subsection{Efficiency of scribble forms}
\label{sec-exp-scribble}


In this study, we first compared four scribble forms to illustrate the efficacy of randomly annotated scribbles for supervision.
Denoting the number of annotated pixels using a manual and skeleton-wise scribble form as $n$, we generated other scribble forms with the same annotated ratios for a fair comparison.
Then, we studied the performance of segmentation with respect to the number of pixels annotated using a random and wide range scribble form, by setting the number of annotated pixels to different times of $n$.
Finally, we further explored variants of random walk annotations to show the importance of wide range in the random distribution of scribbles. 

We applied two segmentation tasks, \ie regular structure segmentation of the cardiac ventricles on ACDC dataset and irregular segmentation of myocardial pathologies using MyoPS dataset. 
To compare the supervision of scribble forms directly, we trained all models with partial cross entropy (PCE) loss calculated for annotated pixels from scribbles.
All experiment results were reported on the test set.

\subsubsection{Scribble forms}  \label{sec:exp-scribbleform}
One can measure the efforts of scribble annotations from two perspectives, \ie number of manual draws to generate scribble annotations ($N_\text{scribble}$) and number of annotated pixels ($N_\text{pix}$).
Given the certain amount of efforts, we designed four forms following different generation procedures, \ie (1) Skeleton, (2) Random walk, (3) Directed random walk (DirRandomWalk), (4) Points, and compared the segmentation performance of models trained using such scribble annotations for supervision.  
The details of scribble forms are described bellow.

\textbf{Skeleton} indicates the widely adopted scribble form by a rater, who approximately outlines the shape of each label class within the segmentation mask. For a segmentation task with $k$ label classes, including the background, one needs $k$ manual draws (scribbles) for a training image. 
For ACDC dataset, we adopted the manual annotated skeleton scribble released by~\cite{9389796}; while for pathologies in MyoPS dataset, we generated the skeleton scribbles automatically using the skeletonization algorithm~\cite{rajchl2017employing}. We refer the reader to Appendix B of the supplementary material for generation details.

\textbf{Random walk} starts from a random point within the segmentation mask.
Then, the annotation moves along a random direction of image lattice within the segmentation mask, with a given step length ($l$ by default set to 1).
We repeated such moves until the ratio or number of annotated pixels reached a threshold ($n$).

\textbf{Directed random walk}, DirRandomWork for short, is the random walk with momentum. 
The scribble generated by Random walk tends to cluster within a local area of the radius $\sqrt{r}$ given $r$-step walks. 
To achieve wide range distribution  without manually setting the step length ($l$), we therefore adopted this directed random walk,
which prefers moving along the same direction to the previous step. 
If the next point does not lie in the segmentation mask, we changed the walking direction to be along the smallest angle to the previous one.

\textbf{Points} scribble form refers to an ideal form, which randomly samples annotated pixels within the segmentation mask. 
However, it is difficult to generate such scribble annotation in practice, due to the huge number of manual draws which equals the number of annotated pixels, \ie $N_\text{scribble}=N_\text{pix}$. Therefore, we considered this form as the upper bound of scribble supervision under the same ratio of annotated pixels.

\begin{table*}[!t]
\caption{Results and comparisons of regular structure segmentation on ACDC dataset. These models were evaluated on the test set.
}\label{tab5}
\centering
\resizebox{\textwidth}{!}{
\begin{tabular}{ccccccccc}
\hline
\multirow{2}{*}[-4pt]{Methods}&\multicolumn{4}{c|}{Dice}&\multicolumn{4}{c}{HD (mm)}\\
\cmidrule(lr){2-5}\cmidrule(lr){6-9}
&LV & MYO & RV &\multicolumn{1}{c|}{Avg} & LV & MYO & RV & Avg \\
\hline
\multicolumn{1}{c|}{PCE}&.805$\pm$.145&.737$\pm$.095&.769$\pm$.128&\multicolumn{1}{c|}{.770$\pm$.126}&62.55$\pm$36.04&68.30$\pm$27.77&59.62$\pm$42.62&63.40$\pm$35.76\\
\multicolumn{1}{c|}{WSL4~\cite{luo2022scribble}}&.835$\pm$.164&\textbf{.825$\pm$.032}&.787$\pm$.191&\multicolumn{1}{c|}{.792$\pm$.166}&\underline{16.48$\pm$16.01}&24.48$\pm$22.74&18.21$\pm$11.30&19.72$\pm$17.67\\
\multicolumn{1}{c|}{GatedCRF~\cite{Obukhov2019GatedCL}}&.846$\pm$.157&.744$\pm$.108&.822$\pm$.111&\multicolumn{1}{c|}{.804$\pm$.135}&37.38$\pm$46.37&22.30$\pm$15.72&20.88$\pm$11.85&26.85$\pm$30.03\\
\multicolumn{1}{c|}{MAAG~\cite{9389796}}&.879&\underline{.817}&.752&\multicolumn{1}{c|}{.816}&25.23&26.83&22.73&24.93\\
\multicolumn{1}{c|}{CVIR~\cite{garg2021mixture}}&.866$\pm$.127&.797$\pm$.102&.737$\pm$.130&\multicolumn{1}{c|}{.800$\pm$.130}&47.51$\pm$50.82&\underline{10.70$\pm$8.39}&\underline{14.39$\pm$9.00}&.24.20$\pm$34.17\\	\multicolumn{1}{c|}{nnPU~\cite{NIPS2017_7cce53cf}}&.862$\pm$.134&.792$\pm$.124&.829$\pm$.102&\multicolumn{1}{c|}{.828$\pm$.123}&67.28$\pm$48.60&18.60$\pm$17.93&14.64$\pm$8.39&33.51$\pm$38.43\\		\multicolumn{1}{c|}{CycleMix~\cite{zhang2022cyclemix}}&.876$\pm$.096&.794$\pm$.083&.829$\pm$.099&\multicolumn{1}{c|}{.833$\pm$.098}&16.60$\pm$19.90&18.04$\pm$17.78&19.09$\pm$21.44&\underline{17.91$\pm$19.57}\\
\multicolumn{1}{c|}{ShapePU~\cite{zhang2022shapepu}}&\underline{.885$\pm$.103}&.806$\pm$.096&\underline{.851$\pm$.089}&\multicolumn{1}{c|}{\underline{.848$\pm$.100}}&20.17$\pm$22.40&41.81$\pm$33.40&20.06$\pm$26.43&27.35$\pm$29.33\\
\multicolumn{1}{c|}{ZScribbleSeg}&\textbf{.900$\pm$.065}&\textbf{.825$\pm$.069}&\textbf{.862$\pm$.102}&\multicolumn{1}{c|}{\textbf{.862$\pm$.086}}&\textbf{7.69$\pm$6.94}&\textbf{8.93$\pm$6.40}&\textbf{12.74$\pm$12.48}&\textbf{9.79$\pm$9.19}\\
\hdashline
\multicolumn{1}{c|}{FullSupUNet}&.882$\pm$.123&.824$\pm$.099&.856$\pm$.112&\multicolumn{1}{c|}{.854$\pm$.113}&11.94$\pm$13.58&12.65$\pm$12.52&14.82$\pm$9.69&13.14$\pm$11.97\\
\hline
\end{tabular}}
\end{table*}

\begin{figure*}[!t]
\centering
\includegraphics[width = \textwidth]{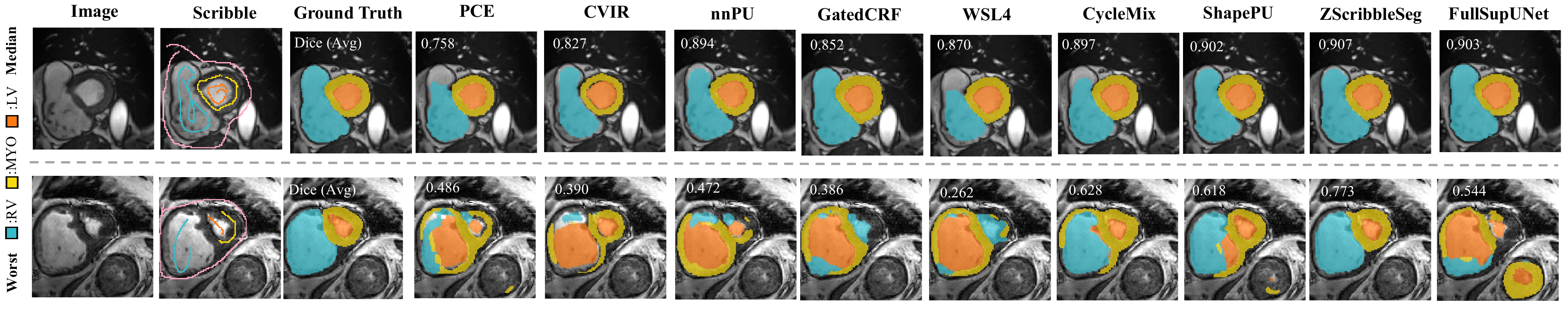} 
\caption{Visualization of cardiac segmentation on ACDC dataset. 
The two slices were from the median and the worst cases by the average Dice scores of all compared methods.}
\label{fig:ACDC-sota}
\end{figure*}

\subsubsection{Results} 

Given the same amount of annotated pixels, we show the effect of different scribble forms on regular structures (ACDC) and irregular objects (MyoPS).
As \zxhreftb{tab1} illustrates, when the four scribble forms had the same number of annotated pixels $N_\text{pix}$, Points achieved the best Dice scores on both of the structural segmentation and irregular segmentation tasks, thanks to the effects of randomness and wide range distribution of scribbles. 
However, when we limited the efforts of manual draws to be the same, DirRandomWalk became more favorable, as the scribble form of Points could be impractical. 
Furthermore, Skeleton scribble was illustrated to be the least efficient form, particularly the segmentation network trained on such dataset performed poorly on the irregular object segmentation task. 
This was probably due to the fact that when the target was difficult to outline, Skeleton form could fail to portray the entire segmentation, leading to poor performance or even a failure in training the segmentation networks. 
On the contrary, randomly distributed scribble forms, such as Random walk and DirRandomWalk, demonstrated their superiority, particularly on the irregular object segmentation with remarkable improvements on average Dice over Skeleton of $24.1\%$ and $30.7\%$, respectively.

\textbf{Number of annotated points:} 
By varying the number of annotated pixels ($N_{\text{pix}}$), we validated the influence of annotated proportions on scribble supervised segmentation.
As shown in \zxhreffig{fig:exp:scribbleform} (a) and  (c), the model performance tended to be improved as $N_{\text{pix}}$ increases, indicating that model training benefited from larger proportion of annotated pixels.
One can observe from \zxhreffig{fig:exp:scribbleform} (a) that the segmentation performance started converging when $N_{\text{pix}}$ reached $2n$. By contrast, for the more difficult segmentation task on irregular objects, as \zxhreffig{fig:exp:scribbleform} (c) illustrates, the model performance converged after    $N_{\text{pix}}\ge 4n$.


\textbf{Wide-ranged distribution:}  
We further investigated the influence of wide range distribution of scribbles, by training networks with varying step length $l$ in Random walk.
As the step length increases, the label distribution range of Random walk gradually expanded.
From \zxhreffig{fig:exp:scribbleform} (b) and (d), one can see that the segmentation performance of average Dice scores was improved as the step length increased, and the performance gradually converged to that of DirRandomWalk.
This confirmed that the widely distributed scribbles were better to provide finer supervision under the same number of draws and annotated pixels.

\subsection{Ablation study}\label{sec-exp-ablation}
We studied the effectiveness of the proposed strategies in modeling efficient scribbles and prior regularization for ZScribbleNet.
We used the ACDC dataset and the expert-made scribble annotations released by \cite{9389796}, and evaluated the model performance on the \textit{validation set}.
We compared six ablated models which were trained  with or without the usage of modeling efficient scribbles (denoted as Efficiency),
and with different combinations of the four loss functions, \ie
the partial cross entropy ($\mathcal{L}_{\text{pce}}$), the global consistency loss ($\mathcal{L}_{\text{global}}$) in Eq.(\ref{4.1eq4}), the spatial prior loss ($\mathcal{L}_{\text{spatial}}$) in Eq.(\ref{eq19}), and the shape prior loss ($\mathcal{L}_{\text{shape}}$) in Eq.(\ref{4.2eq1}).

\zxhreftb{tab3} presents the results. 
When model~\#2 adopted the proposed supervision augmentation to model efficient scribbles (indicated by the column of Efficiency), its performance was improved compared to model~\#1, as one can see from their average Dice scores (0.848 vs. 0.813) and average HDs (65.35 mm vs. 69.58 mm).
This demonstrated the benefits of model training from the augmented supervision.
When combining the supervision augmentation with the global consistency loss ($\mathcal{L}_{\text{global}}$), leading to model~\#4, the performance was further boosted with remarkable improvements, namely 4.3\% gain in Dice (0.891  vs. 0.848) and over 45 mm error reduction in HD (19.50 mm vs. 65.35 mm).
Alternatively, when leveraging inter connectivity via the shape regularization loss ($\mathcal{L}_{\text{shape}}$), model~\#3 obtained an overwhelming improvement in HD, which was reduced from 69.58 mm to only 15.16 mm compared to model~\#1. This indicated the results were with much less noisy and outlier segmentation. 
We then further investigated the advantage of spatial prior ($\mathcal{L}_{\text{spatial}}$) in training ZScribbleNet.
One can see from the result of model~\#5 that it achieved the most evident gain in terms of Dice results, with an improvement of 8.1\% (0.894 vs. 0.813) by solely including one extra loss. 
Finally, our ZScribbleSeg (model~\#6) achieved the best performance with average Dice of 0.899 and HD of $8.70$ mm.
This indicated that the combination of efficient scribbles and priors endowed the algorithm with substantial supervision and prior knowledge for scribble-supervised segmentation.
    
\begin{table*}[!t]
\caption{Results and comparisons of regular structure segmentation on pathology enhanced images (LGE CMR) using MSCMRseg dataset.} \label{tab6}
\centering
\resizebox{1\textwidth}{!}{
\begin{tabular}{ccccccccc}
\hline
\multirow{2}{*}[-4pt]{Methods}&\multicolumn{4}{c|}{Dice}&\multicolumn{4}{c}{HD (mm)}\\
\cmidrule(lr){2-5}\cmidrule(lr){6-9}
&LV & MYO & RV &\multicolumn{1}{c|}{Avg} & LV & MYO & RV & Avg \\
\hline
\multicolumn{1}{c|}{PCE}&.514$\pm$.078&.582$\pm$.067&.058$\pm$.023&\multicolumn{1}{c|}{.385$\pm$.243}&259.4$\pm$14.19&228.1$\pm$21.36&257.4$\pm$12.43&248.3$\pm$21.63\\
\multicolumn{1}{c|}{WSL4~\cite{luo2022scribble}}&.902$\pm$.040&.815$\pm$.033&.828$\pm$.101&\multicolumn{1}{c|}{.848$\pm$.076}&55.95$\pm$4.88&42.07$\pm$13.48&\textbf{32.08$\pm$6.57}&43.37$\pm$31.04\\
\multicolumn{1}{c|}{GatedCRF~\cite{Obukhov2019GatedCL}}&\underline{.917$\pm$.044}&\underline{.825$\pm$.032}&\underline{.848$\pm$.073}&\multicolumn{1}{c|}{\underline{.863$\pm$.066}}&\underline{25.72$\pm$4.37}&37.92$\pm$5.10&\underline{32.83$\pm$5.59}&\underline{32.16$\pm$7.11}\\
\multicolumn{1}{c|}{CVIR~\cite{garg2021mixture}}&.331$\pm$.076&.371$\pm$.088&.404$\pm$.110&\multicolumn{1}{c|}{.368$\pm$.095}&259.2$\pm$14.23&243.0$\pm$13.76&180.9$\pm$55.44&227.7$\pm$47.63\\
\multicolumn{1}{c|}{nnPU~\cite{NIPS2017_7cce53cf}}&.341$\pm$.067&.538$\pm$.081&.432$\pm$.100&\multicolumn{1}{c|}{.437$\pm$.115}&259.4$\pm$14.19&201.6$\pm$66.98&199.7$\pm$57.50&220.2$\pm$57.70\\		\multicolumn{1}{c|}{CycleMix~\cite{zhang2022cyclemix}}&.748$\pm$.064&.730$\pm$.047&.835$\pm$.041&\multicolumn{1}{c|}{.771$\pm$.069}&224.59$\pm$35.27&\underline{28.26$\pm$20.77}&73.36$\pm$51.39&108.74$\pm$92.65\\ \multicolumn{1}{c|}{ShapePU~\cite{zhang2022shapepu}}&.880$\pm$.046&.785$\pm$.080&.833$\pm$.087&\multicolumn{1}{c|}{.833$\pm$.082}&178.02$\pm$50.93&178.05$\pm$25.39&189.35$\pm$55.78&181.81$\pm$45.27\\
\multicolumn{1}{c|}{ZScribbleSeg}&\textbf{.922$\pm$.039}&\textbf{.834$\pm$.039}&\textbf{.854$\pm$.055}&\multicolumn{1}{c|}{\textbf{.870$\pm$.058}}&\textbf{12.10$\pm$14.70}&\textbf{16.52$\pm$19.14}&51.03$\pm$39.27&\textbf{26.55$\pm$31.39}\\
\hdashline
\multicolumn{1}{c|}{FullSupUNet}&.909$\pm$.049&.821$\pm$.054&.826$\pm$.087&\multicolumn{1}{c|}{.852$\pm$.076}&10.02$\pm$12.36&11.89$\pm$11.34&56.91$\pm$41.99&26.27$\pm$33.63\\
\hline
\end{tabular}}
\end{table*}  

\begin{figure*}[!t]
\centering
\includegraphics[width = \textwidth]{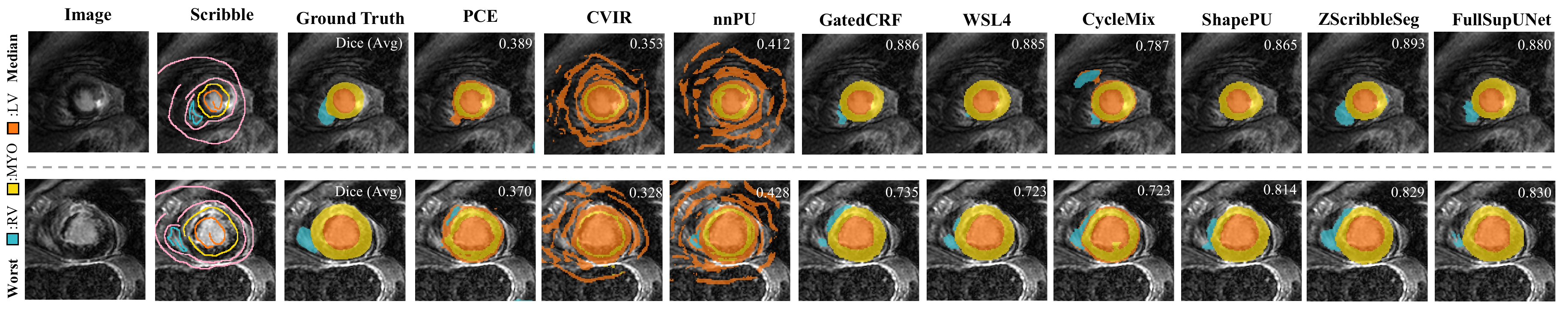}
\caption{Visualization of cardiac segmentation on LGE CMR using MSCMRseg dataset. The two slices were from the median and the worst cases by the average Dice scores of all compared methods.}
\label{fig:mscmr-sota}
\end{figure*}

\subsection{Performance and Comparisons} 
\label{sec-exp-sota}
    
We conducted experiments over the \emph{four} segmentation tasks stated in Section~\ref{sec-exp-materials}.
\textbf{(1)} For the structural segmentation of cardiac ventricles from ACDC dataset, we used the expert-made scribbles released by \cite{9389796}.
\textbf{(2)} For the cardiac structural segmentation from pathology enhanced imaging (MSCMRseg) dataset,  we used the manually annotated scribbles  released by~\cite{zhang2022cyclemix}. 
\textbf{(3)} For the irregular myocardial pathology segmentation from MyoPS dataset, we first adopted the standard skeletonization algorithm for the simulated scribble annotation of pathologies \cite{rajchl2017employing}. Then, we manually annotated skeleton scribbles for the structures of LV, Myo, RV and background. 
\textbf{(4)} For the human pose segmentation from PPSS dataset, we adopted the scribble annotations generated by the standard skeletonization algorithm \cite{rajchl2017employing}.

We compared ZScribbleSeg with eight to nine methods.
We first implemented the PCE loss ($\mathcal{L}_{\text{pce}}$) as a baseline method (referred to PCE).
Then, we implemented four state-of-the-art (SOTA) scribble supervised segmentation methods, \ie WSL4~\cite{luo2022scribble}, GatedCRF~\cite{Obukhov2019GatedCL}, CycleMix~\cite{zhang2022cyclemix}, and ShapePU~\cite{zhang2022shapepu} to run the same experiments.
We  cited the ACDC and PPSS results reported in \cite{9389796} for the MAAG method, which is also a SOTA method for this task. 
Furthermore, we adopted two semi-supervised SOTA methods based on positive unlabeled learning, \ie  CVIR~\cite{garg2021mixture} and nnPU~\cite{NIPS2017_7cce53cf}, and re-implemented to adapt them for the scribble-supervised segmentation tasks.
For more details of adaptation, the readers are referred to Appendix C of the supplementary material.
Finally, we trained UNet with full annotations as a baseline of fully-supervised approach (referred to as FullSupUNet).
Note that the post-processing steps of all experiments were removed for a fair comparison. 
    

\subsubsection{Structure segmentation from anatomical images} \label{se-exp-sota:1}
\zxhreftb{tab5} presents the Dice and HD results of 10 approaches for regular structure segmentation of cardiac ventricles from ACDC dataset. 
One can observe that ZScribbleSeg achieved average Dice of 0.862 and HD of 9.79 mm, outperforming the other scribble-supervised methods evidently.
The quantitative results of ZScribbleSeg were comparable to (or slightly better than) that of the fully supervised method (FullsupUNet) whose average Dice and HD are 0.854 and 13.14 mm, respectively.

Particularly, the HD results of ZScribbleSeg (9.79 mm) and FullSupUNet (13.14 mm) were evidently much better than the other methods. 
Note that HD is highly sensitive to the noisy and outlier segmentation results, which are commonly seen when the supervision of global shape information is not sufficient. 
The results indicate the proposed efficient scribble modeling and prior regularization were able to alleviate the problem of inadequate supervision and incomplete shape information from training images with scribble annotations.
Finally, \zxhreffig{fig:ACDC-sota} visualizes two typical cases (median and worst) for illustration.

\begin{table*}[!t]
\caption{Results and comparisons of irregular segmentation of myocardial pathologies on MyoPS dataset.}\label{tab7}
\centering
\resizebox{0.9\textwidth}{!}{
\begin{tabular}{ccccccc}
\hline
\multirow{2}{*}[-4pt]{Methods}&\multicolumn{3}{c|}{Dice}&\multicolumn{3}{c}{HD (mm)}\\
\cmidrule(lr){2-4}\cmidrule(lr){5-7}
& Scar & Edema &\multicolumn{1}{c|}{Avg} & Scar & Edema & Avg \\
\hline
\multicolumn{1}{c|}{PCE}&0.504$\pm$0.213&0.057$\pm$0.022&\multicolumn{1}{c|}{0.281$\pm$0.271}&82.68$\pm$33.95&147.61$\pm$20.59&115.15$\pm$43.00\\
\multicolumn{1}{c|}{WSL4~\cite{luo2022scribble}}&0.031$\pm$0.029&0.106$\pm$0.033&\multicolumn{1}{c|}{0.069$\pm$0.049}&172.37$\pm$45.13&170.05$\pm$20.44&171.20$\pm$34.60\\
\multicolumn{1}{c|}{GatedCRF~\cite{Obukhov2019GatedCL}}&0.020$\pm$0.013&0.042$\pm$0.020&\multicolumn{1}{c|}{0.031$\pm$0.019}&173.60$\pm$44.98&170.10$\pm$20.44&171.8$\pm$34.53\\
\multicolumn{1}{c|}{CVIR~\cite{garg2021mixture}}&0.505$\pm$0.214&0.080$\pm$0.031&\multicolumn{1}{c|}{0.293$\pm$0.263}&61.59$\pm$32.09&125.27$\pm$20.83&93.43$\pm$41.86\\	\multicolumn{1}{c|}{nnPU~\cite{NIPS2017_7cce53cf}}&0.530$\pm$0.241&0.085$\pm$0.035&\multicolumn{1}{c|}{0.308$\pm$0.282}&\underline{48.88$\pm$23.55}&125.27$\pm$20.83&87.07$\pm$44.47\\		\multicolumn{1}{c|}{CycleMix~\cite{zhang2022cyclemix}}&0.550$\pm$0.237&\underline{0.626$\pm$0.124}&\multicolumn{1}{c|}{\underline{0.588$\pm$0.191}}&65.64$\pm$42.81&81.97$\pm$40.87&73.81$\pm$42.13\\
\multicolumn{1}{c|}{ShapePU~\cite{zhang2022shapepu}}&\underline{0.558$\pm$0.237}&0.615$\pm$0.144&\multicolumn{1}{c|}{0.587$\pm$0.205}&57.33$\pm$31.58&\underline{53.00$\pm$31.42}&\underline{55.16$\pm$31.17}\\
\multicolumn{1}{c|}{ZScribbleSeg}&\textbf{0.596$\pm$0.237}&\textbf{0.676$\pm$0.113}&\multicolumn{1}{c|}{\textbf{0.636$\pm$0.188}}&\textbf{46.73$\pm$20.04}&\textbf{47.05$\pm$24.30}&\textbf{46.89$\pm$21.98}\\
\hdashline
\multicolumn{1}{c|}{FullSupUNet}&0.607$\pm$0.253&0.659$\pm$0.135&\multicolumn{1}{c|}{0.633$\pm$0.202}&55.35$\pm$35.73&63.53$\pm$33.15&59.44$\pm$34.27\\
\hline
\end{tabular}}
\end{table*}

\begin{figure*}[!t]
\centering
\includegraphics[width = 1\textwidth]{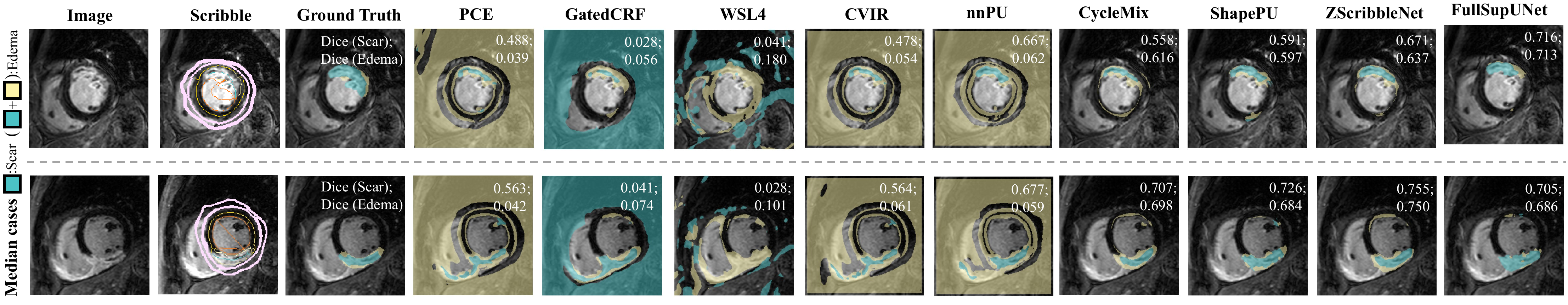}
\caption{Visualization of irregular segmentation of myocardial pathologies on MyoPS dataset. The two slices were from the median cases by average Dice scores of edema or scar segmentation of all compared methods.
}
\label{fig:myops-sota}
\end{figure*}

\subsubsection{Structure segmentation from pathology enhanced images} 
The anatomical segmentation from pathology enhanced images, \ie LGE CMR of MSCMRseg dataset, was a more challenging task compared to that of ACDC dataset.
This is because MSCMRseg was a smaller dataset (e.g.: 25 vs. 70 training subjects), and the image quality and appearance pattern of LGE CMR could be worse and more complex.

\zxhreftb{tab6} provides the quantitative results, and \zxhreffig{fig:mscmr-sota} visualizes two special examples (median and worst) for demonstration.
ZScribbleSeg achieved promising performance and better Dice and HD results than the other SOTA methods for scribble supervised segmentation. 
Notice that for this particular challenging task, the two general semi-supervised  segmentation methods, \ie CVIR and nnPU, could not work properly, which was confirmed by the two failed segmentation examples visualized in \zxhreffig{fig:mscmr-sota}.

Finally, similar to the results in previous study (Section~\ref{se-exp-sota:1}), ZScribbleSeg and FullSupUNet could achieve less noisy segmentation, affirmed by the remarkable better HD results in \zxhreftb{tab6}.
Hence, we second to the conclusion that the proposed ZScribbleNet received greatly augmented supervision and global shape information via the proposed efficient scribble modeling and prior regularization.

\subsubsection{Irregular segmentation}
For segmentation of objects with heterogeneous shape features, it becomes particularly challenging to learn the accurate shape information for inference. 
We evaluated ZScribbleSeg on such challenging task of irregular segmentation using myocardial pathology segmentation (MyoPS), where \textit{we removed the shape regularization term $\mathcal{L}_{\text{shape}}$ due to the nature of pathologies lacking such property}.

\zxhreftb{tab7} shows the performance in detail, and \zxhreffig{fig:myops-sota} visualizes two typical cases, \ie median cases by average Dice scores of edema and scar segmentation, respectively.
One can find that the advantages of the proposed methodologies were demonstrated evidently in such challenging task, as the performance gains, either in terms of Dice or HD, were significant from CycleMix, ShapePU and finally to ZScribbleSeg compared to PCE, WSL4, GatedCRF, CVIR and nnPU ($p\!\!<\!\!0.001$).
In fact, the scribble-supervised segmentation of edema by the compared five methods were failed, and so were the segmentation of scar for WSL4 and GatedCRF. This is illustrated in the visualized examples in \zxhreffig{fig:myops-sota}. 
Although WSL4 and GatedCRF worked well, with scribble supervision, in the above two regular structure segmentation tasks, they suffered severely from noisy labels due to their dependence of training on pseudo labels, which leads to the failure of model training.  
Furthermore, due to the similar texture between edema and surrounding tissues in all imaging modalities, it could be extremely difficult to segment such pathology relying solely on training images without robust estimation and regularization of class mixture ratios. One can see from the result that this failed all the five compared methods in edema segmentation.
By contrast, ShapePU and ZScribbleSeg succeeded in this task thanks to their own methods of estimating class prior $\bm{\pi}$ and applying spatial regularization, which is affirmed by the fact that they both achieved good HDs comparable to that of FullSupUNet for scar and edema segmentation. 
Notice that CycleMix did not illustrate such good performance in terms of HDs, but it achieved comparable good Dice scores thanks to the adoption of supervision augmentation.  
\begin{figure*}[!thb]
\centering
\includegraphics[width =\textwidth]{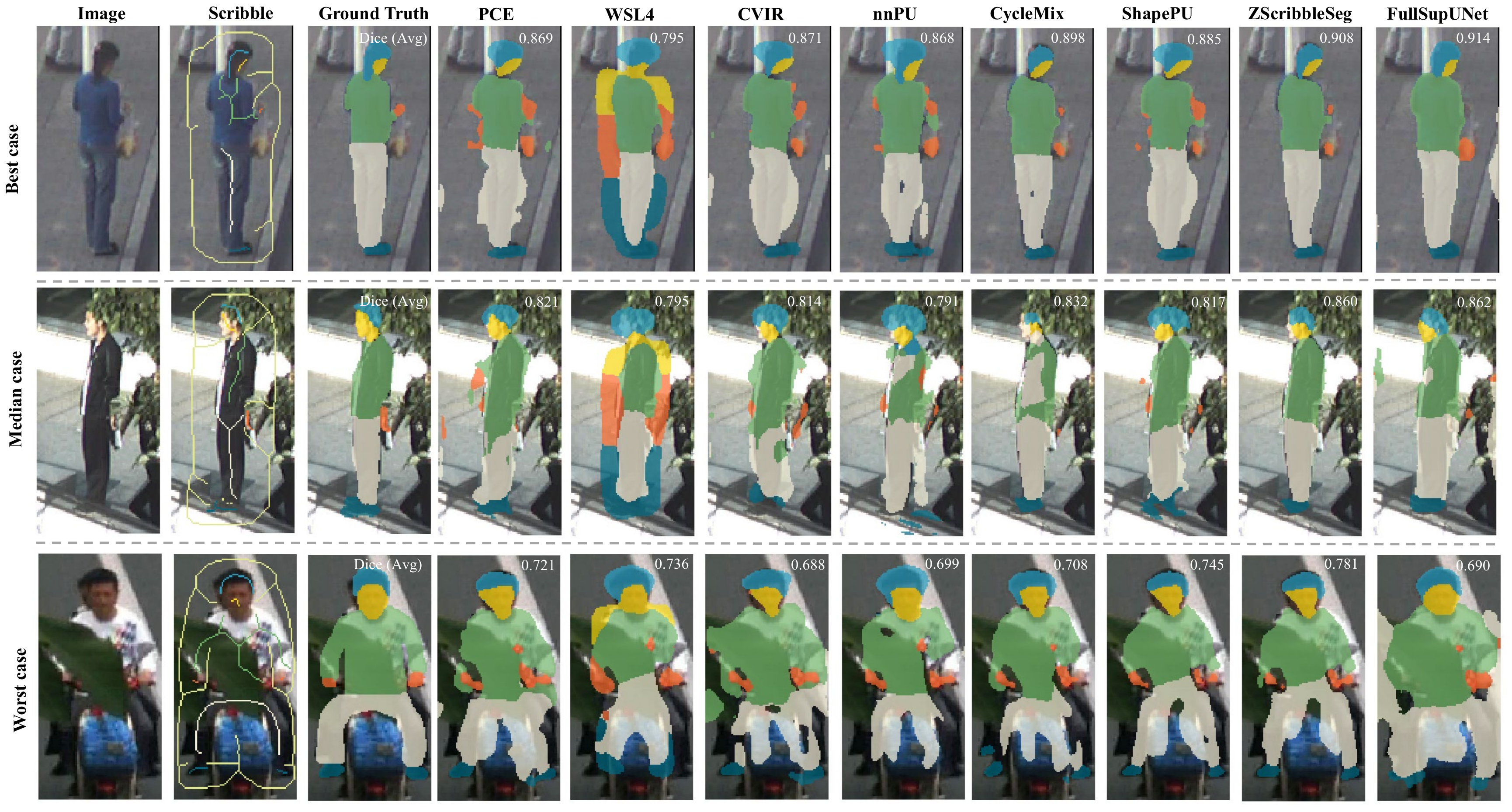}
\caption{Visualization of results on PPSS dataset.  The selected subjects were the best, median and  worst cases by the average Dice scores of all compared methods.}
\label{fig:PPSS-sota}
\end{figure*}

\subsubsection{Segmentation from natural scenes}
\begin{table}[!t]
\caption{Dice results of the 10 methods on the four datasets. Note that sizes of training sets are given in the brackets.}\label{tab_overview}
\centering
\resizebox{1\linewidth}{!}{
\begin{tabular}{ccccc}
\hline
\multirow{2}{*}{Methods}  
&ACDC & MSCMRseg & MyoPS &PPSS  \\ 
&(70) & (25) & (20) &(2828) \\
\hline
\multicolumn{1}{c|}{PCE}&.770$\pm$.126&.385$\pm$.243&.281$\pm$.271&.805$\pm$.063\\
\multicolumn{1}{c|}{WSL4~\cite{luo2022scribble}}&.792$\pm$.166&\underline{.848$\pm$.076}&-&.762$\pm$.045\\
\multicolumn{1}{c|}{GatedCRF~\cite{Obukhov2019GatedCL}}&.804$\pm$.135&.825$\pm$.032&-&-\\
\multicolumn{1}{c|}{MAAG~\cite{9389796}}&.816&-&-&.746\\
\multicolumn{1}{c|}{CVIR~\cite{garg2021mixture}}&.800$\pm$.130&.368$\pm$.095&.293$\pm$.263&.809$\pm$.054\\
\multicolumn{1}{c|}{nnPU~\cite{NIPS2017_7cce53cf}}&.828$\pm$.123&.437$\pm$.115&.308$\pm$.282&.794$\pm$.055\\		
\multicolumn{1}{c|}{CycleMix~\cite{zhang2022cyclemix}}&.833$\pm$.098&.771$\pm$.069&\underline{.588$\pm$.191}&\underline{.835$\pm$.050}\\ 
\multicolumn{1}{c|}{ShapePU~\cite{zhang2022shapepu}}&\underline{.848$\pm$.100}&.833$\pm$.082&.587$\pm$.205&.823$\pm$.055\\
\multicolumn{1}{c|}{ZScribbleSeg}&\bf{.862$\pm$.086}&\bf{.870$\pm$.058}&\bf{.636$\pm$.188}&\bf{.838$\pm$.050}\\
\hdashline
\multicolumn{1}{c|}{FullSupUNet}&.854$\pm$.113&.852$\pm$.076&.633$\pm$.202&.843$\pm$.071\\
\hline
\end{tabular}}
\end{table}

We further validated the broad utility of ZScribbleSeg on the human pose segmentation task of natural scene images.
We applied all the methods on the PPSS dataset, which consists of pedestrian images with occlusions, generated by different cameras with different resolutions.
 
\zxhreftb{tab_overview} presents the details, together with the summarized results from previous three studies, \ie ACDC, MSCMRseg and MyoPS.
Similar to the three medical image segmentation tasks, the model of ZScribbleSeg generalized well to this 3-channel colored natural image segmentation task, with the performance comparable to FullSupUNet and Dice accuracy setting new state of the art for scribble supervised segmentation.

\zxhreffig{fig:PPSS-sota} visualizes three special cases, \ie the best, median and the worst cases according to the average Dice by all compared methods.  
One can see from the figures that ZScribbleNet performed robustly and generated realistic segmentation with less noisy results, particularly compared with other scribble supervised methods and the fully supervised one (FullSupUNet).
 
\section{Conclusion} \label{section6}

In this work, we have presented a new framework for scribble-supervised segmentation, \ie ZScribbleSeg, to integrate the efficient scribbles and prior regularization with implementation of a deep neural network (ZScribbleNet).
ZScribbleSeg exploits the principles of "good scribble annotations", and effectively augments the scribble supervision of ZScribbleNet, via mixup-occlusion operations and global consistency regularization. 
Then, we explored to capture the global information by incorporating the prior information, particularly with proposals of spatial prior loss and shape prior loss. 
The spatial prior loss was based on the estimated spatial energy and label class mixture proportions $\bm{\pi}$. 
The former provides a new means to identify the probability of unlabeled pixels belonging to each class without  directly using model predictions; 
and the latter was developed based on a novel estimation method and was aimed to correct the problematic prediction via the regularization of spatial prior loss.

To examine to performance of ZScribbleSeg, we investigated a variety of segmentation tasks, including regular structural segmentation of cardiac ventricles from anatomical imaging data (using ACDC dataset), regular structural segmentation of pathology enhanced imaging data (MSCMRseg), irregular object segmentation from multi-modality imaging (MyoPS), and human pose segmentation from natural scenario (PPSS).
Compared to others approaches, ZScribbleSeg has shown great competence and achieved comparable performance to the fully supervised UNet method. 
Particularly,  thanks to the augmented supervision and prior regularization, ZScribbleSeg performed well and demonstrated reliability and generalizability in the scenarios with small training set (MSCMRseg task) and irregular structure segmentation (MyoPS task), both of which failed several other state-of-the-art approaches. 
%
%
%
\bibliographystyle{splncs04}
\bibliography{citation.bib}

\end{document}